\documentclass[11pt]{article}

\usepackage[preprint]{acl}
\usepackage{subcaption}
\usepackage{graphicx}
\usepackage{times}
\usepackage{latexsym}
\usepackage{enumitem}
\usepackage[T1]{fontenc}
\usepackage{booktabs}
\usepackage{multirow}
\usepackage{colortbl}
\usepackage{xcolor}
\usepackage{pifont} 
\usepackage{bm}
\usepackage[utf8]{inputenc}

\usepackage{microtype}
\usepackage{amsfonts}
\usepackage{booktabs}
\usepackage{multirow}
\usepackage{xcolor}

\usepackage{inconsolata}

\usepackage{graphicx}
\usepackage{amsmath}
\usepackage{booktabs}
\usepackage{multirow}
\usepackage{xcolor}
\usepackage{colortbl} 
\usepackage[table]{xcolor} 
\definecolor{teal}{RGB}{0,128,128} 
\definecolor{purple}{RGB}{128,0,128}
\definecolor{bestclosed}{HTML}{F4CCCC} 
\definecolor{bestopen}{HTML}{CFE2F3}   
\definecolor{deepred}{HTML}{8B0000} 

\title{Thinking with Constructions: A Benchmark and Policy Optimization for Visual-Text Interleaved Geometric Reasoning}
\author{
  \textbf{Haokun Zhao}\textsuperscript{1,$*\dagger$}\thanks{\ \ $^*$These authors contributed equally to this work. $^\dagger$Corresponding author.} \quad
  \textbf{Wanshi Xu}\textsuperscript{2,$*$} \quad
  \textbf{Haidong Yuan}\textsuperscript{3} \quad
  \textbf{Songjun Cao}\textsuperscript{4} \\
  \textbf{Long Ma}\textsuperscript{4,$\dagger$} \quad
  \textbf{Yanghua Xiao}\textsuperscript{1,$\dagger$} \\
  \\ 
  \textsuperscript{1}College of Computer Science and Artificial Intelligence, Fudan University \\
  \textsuperscript{2}School of ECE, Peking University \\
  \textsuperscript{3}School of Software and Microelectronics, Peking University \\
  \textsuperscript{4}Tencent Youtu Lab \\
  \texttt{hkzhao23@m.fudan.edu.cn}, \texttt{shawyh@fudan.edu.cn} \\
  \texttt{\{xwanshi, oseast\}@stu.pku.edu.cn}, \texttt{\{songjuncao, malonema\}@tencent.com}
}

\begin{document}
\maketitle
\begin{abstract}
Geometric reasoning inherently requires \textbf{"thinking with constructions"}—the dynamic manipulation of visual aids to bridge the gap between problem conditions and solutions. However, existing Multimodal Large Language Models (MLLMs) are largely confined to passive inference with static diagrams, lacking the strategic knowledge of when and how to construct effective visual aids. To address this, we present a framework for \textbf{Visual-Text Interleaved Chain-of-Thought}. We first introduce \textbf{GeoAux-Bench}, the first benchmark comprising 4,334 geometry problems that aligns textual construction steps with ground-truth visual updates. Our pilot study reveals two critical insights: (1) interleaved visual-textual aids outperform single-modality counterparts, which cannot losslessly capture geometric synergy; and (2) valid constructions act as entropy reducers, strongly correlating with reduced reasoning perplexity. Building on these findings, we propose \textbf{Action Applicability Policy Optimization (A\textsuperscript{2}PO)}, a reinforcement learning paradigm for mastering strategic construction. A employs Adaptive Reward Shaping to regulate the timing and quality of visual aids via counterfactual sampling to distinguish necessary from redundant constructions. Experiments demonstrate our approach enables MLLMs to leverage selective auxiliary constructions, yielding a 3.51\% gain over strong baselines. Code and data are available on GitHub \footnote{https://anonymous.4open.science/r/GeoAux-5863}.
\end{abstract}

\section{Introduction}

Recent advancements in Large Language Models (LLMs) have demonstrated remarkable proficiency in mathematical reasoning \cite{Shao2024DeepSeekMathPT,Yang2024Qwen25MathTR,Chen2025SeedProver1M}, largely driven by the Chain-of-Thought (CoT) technique \cite{Wei2022ChainOT}. However, geometry problem solving remains a significant hurdle. Unlike algebraic tasks that rely on symbolic manipulation, geometric reasoning is intrinsically multimodal \cite{Lu2021InterGPSIG,Kazemi2023GeomVerseAS}: human experts do not merely read static diagrams; they solve problems by constructing and manipulating visual aids (e.g., drawing auxiliary lines) to bridge the gap between conditions and solutions. This process of \textbf{auxiliary construction} is the quintessential embodiment of \textit{``thinking with images,''} where the visual context dynamically evolves to reveal hidden geometric properties \cite{Chern2025ThinkingWG,Li2025ImagineWR}.

\begin{figure}[t]
  \centering
  \includegraphics[width=0.48\textwidth]{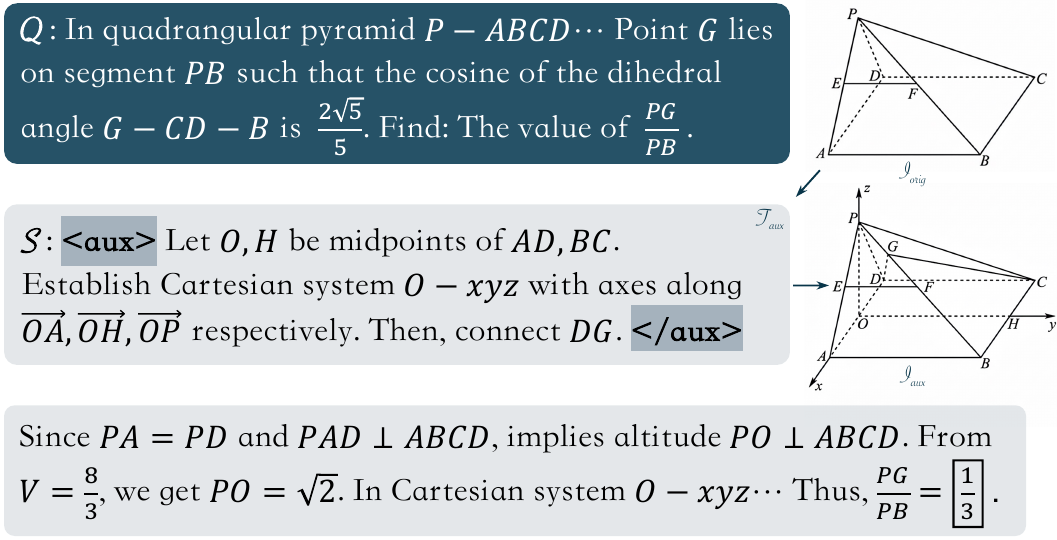}
  \caption{A representative sample from GeoAux-Bench. The solution trajectory is structured in a \textbf{visual-text interleaved format}: the textual auxiliary construction step (e.g., \texttt{<aux>... </aux>}) is explicitly paired with a corresponding auxiliary diagram ($I_{aux}$).  }
  \label{fig:sample}
\end{figure}

To emulate this process, the community has actively explored \textbf{Visual Chain-of-Thought (VCoT)}. However, existing paradigms generally fall into three categories, each facing distinct limitations: (1) \textbf{Agent-based approaches} manipulate geometric code to render diagrams but often rely on ground-truth code inputs, diverging from the natural visual perception of raw images \cite{Hu2024VisualSS,Wang2025VisuoThinkEL}; (2) \textbf{Formal Abstraction} methods convert diagrams into formal languages, acting as a \textit{lossy compression} that strips away visual intuition and risks hallucination \cite{Sharma2024GeoCoderSG,trinh2024solving,Yang2025R1OnevisionAG}; and (3) \textbf{Unified Multimodal Models} attempt to natively generate visual thoughts \cite{Shi2025MathCanvasIV,Li2025ZebraCoTAD}. While promising, even SOTA models like Nano-Banana-Pro \cite{geminiteam2025geminifamilyhighlycapable} suffer from pixel-level structural hallucinations that mislead the reasoning process. Consequently, most MLLMs are confined to a passive static inference mode, unable to update their visual context to match their reasoning steps.

This disconnect highlights a critical gap: models lack the strategic procedural knowledge to employ visual aids effectively—specifically, the decision-making of \textit{when to draw, what to draw, and how to leverage the visualization} for subsequent deduction.
Crucially, such constructions significantly reduce problem-solving difficulty, particularly when diagrams are inherently complex or benefit from intrinsic properties~\cite{Chervonyi2025GoldmedalistPI}.

To address this, we conducted an ablation study by injecting oracle auxiliary aids in different modalities.
Our findings confirm that single-modality auxiliary aids—whether consisting solely of textual instructions or visual diagrams—cannot serve as a lossless substitute for the interleaved provision of construction statements and corresponding images, as they fail to fully encapsulate the synergistic information inherent in the multimodal context.
Furthermore, we observed that this visual feedback correlates strongly with a reduction in reasoning perplexity (PPL), mirroring a \textit{``cognitive epiphany''} where correctly constructed auxiliary lines drastically reduce the uncertainty of the subsequent reasoning trajectory.

Building on these insights, we propose a framework for Visual-Text Interleaved Chain-of-Thought.
We first introduce \textbf{GeoAux-Bench}, a benchmark comprising 4,334 geometry problems and 8,470 diagrams. 
As illustrated in Figure~\ref{fig:sample}, it establishes a precise interleaved mapping where each textual construction ($T_{\text{aux}}$) is explicitly paired with its corresponding ground-truth visual update ($I_{\text{aux}}$).
To effectively leverage this data, we present \textbf{Action Applicability Policy Optimization} (A\textsuperscript{2}PO), a reinforcement learning paradigm designed to master strategic visual construction.
A\textsuperscript{2}PO employs a Tri-Partition Sampling strategy to construct counterfactual reasoning paths (mandatory vs. prohibited). Based on these baselines, we employ Adaptive Reward Shaping to orchestrate the reasoning process:
(1) A \textbf{Timing Reward} to discern the necessity of auxiliary lines;
complemented by (2) a \textbf{Quality Reward}, grounded in reasoning perplexity, to ensure constructions genuinely simplify the solution path.
During inference, we implement Visual Re-prompting to dynamically inject auxiliary diagrams, enabling the model to reason in a truly interleaved manner.

In summary, our contributions are as follows:
\begin{enumerate}[noitemsep,leftmargin=*]
    \item We present \textbf{GeoAux-Bench}, the first geometry benchmark to explicitly associate textual auxiliary construction steps with corresponding auxiliary diagrams.
    
    \item We provide empirical evidence that interleaved visual-textual auxiliary representations significantly outperform single-modality counterparts by up to \textbf{1.97\%}. We further demonstrate that high-quality auxiliary constructions act as an \textbf{entropy reducer}, narrowing the solution search space and lowering reasoning uncertainty.

    \item We propose \textbf{A\textsuperscript{2}PO}, a reinforcement learning paradigm utilizing \textbf{Adaptive Reward Shaping}. By strictly regulating the \textit{timing} and evaluating the \textit{quality} of visual aids, A\textsuperscript{2}PO achieves a maximum performance gain of \textbf{3.51\%} over GRPO and unconditional reinforcement strategies.
\end{enumerate}
\section{Related Work}

\subsection{Benchmarks for Multimodal Mathematical Reasoning}
Visual-mathematical reasoning benchmarks, from foundational datasets \cite{Lu2021InterGPSIG,Lu2022LearnTE} to recent suites like MMMU \cite{Yue2023MMMUAM} and MathVista \cite{Lu2023MathVistaEM}, have advanced MLLMs \cite{Wang2024MeasuringMM,Zhang2024MathVerseDY,Kazemi2023GeomVerseAS}. However, they rely on static pairs, lacking step-by-step visual demonstrations for dynamic reasoning. While \citet{Shi2025MathCanvasIV} explored interleaved formats, such data remains scarce. GeoAux-Bench bridges this gap via explicit text-visual alignment to provide dense supervision.

\subsection{Visual Chain-of-Thought (VCoT)}
Textual Chain-of-Thought \cite{Wei2022ChainOT,Fang2025FLUXReason6MP,Fang2025GoTUR} dominates symbolic reasoning but fails to capture spatial dynamics for geometry. Visual CoT addresses this by integrating visual synthesis into deduction, with two primary streams: (1) \textbf{Tool-augmented approaches} \cite{Hu2024VisualSS,Wang2025VisuoThinkEL,Zheng2025DeepEyesI} suffer from procedural rigidity, treating diagramming as a static step rather than a fluid cognitive strategy; (2) \textbf{Intrinsic generation models} \cite{Shi2025MathCanvasIV,Li2025ZebraCoTAD} are prone to structural hallucinations and lack geometric precision. Our framework synthesizes these paradigms via Visual Re-prompting, combining symbolic precision with dynamic reasoning feedback.

\subsection{Reinforcement Learning for Reasoning}
Reinforcement learning is central to reasoning breakthroughs \cite{DeepSeekAI2025DeepSeekR1IR}. Algorithms like GRPO \cite{Shao2024DeepSeekMathPT}, DAPO \cite{Yu2025DAPOAO}, and VAPO \cite{Yue2025VAPOEA} excel in text, vision \cite{Meng2025MMEurekaET}, and tool-use \cite{Li2025ToRLST,hong2025deepeyesv2agenticmultimodalmodel}, but applying RL to geometry is non-trivial—intermediate visual constructions lack verifiable outcomes. Existing methods like GeometryZero \cite{Wang2025GeometryZeroIG} optimize timing via tool priors \cite{Li2025ToRLST} but overlook reasoning efficacy. Our A\textsuperscript{2}PO addresses this via Adaptive Reward Shaping, using contrastive sampling for timing and reasoning perplexity for utility assessment, ensuring constructions actively facilitate solutions.


\section{The GeoAux Benchmark and Reasoning Dynamics}
\label{sec:benchmark_and_dynamics}
In this section, we introduce \textbf{GeoAux}, a benchmark tailored for \textbf{Interleaved-Modal Chain-of-Thought (VCoT)}. To capture the dynamic nature of geometric reasoning, GeoAux explicitly aligns textual auxiliary instructions with their corresponding visual updates ($T_{aux} \leftrightarrow I_{aux}$). We first detail the benchmark construction pipeline, followed by a pilot study that quantitatively validates the cognitive synergy between these modalities.

\subsection{GeoAux-Bench Construction}
\label{sec:benchmark_construction}

To foster research into active visual-textual reasoning, we construct \textbf{GeoAux-Bench}, consisting of two subsets: the expert-annotated \textbf{GeoAux-Core} and the adapted \textbf{GeoAux-Canvas}.

\paragraph{GeoAux-Core.}
We curated geometric problems predominantly requiring auxiliary constructions from secondary school curricula and Olympiad competitions (post-January 2024) to minimize data contamination.
Crucially, we restructure the solution trajectory into an interleaved format. We introduce a dedicated token pair \texttt{<aux>}...\texttt{</aux>} to encapsulate the Auxiliary Construction Instruction ($T_{aux}$). The closing tag \texttt{</aux>} explicitly marks the insertion point for the corresponding Auxiliary Diagram ($I_{aux}$). This structure acts as a transition operator mapping the initial visual state ($I_{orig}$) to an updated state ($I_{aux}$):

\begin{equation}
    \mathcal{M}: (I_{orig}, T_{aux}) \rightarrow I_{aux}
\end{equation}

Here, $T_{aux}$ contains explicit directives (e.g., ``\textit{Connect $AB$}''), while $I_{aux}$ renders these elements visually. A representative sample is shown in Figure~\ref{fig:sample}.
The subset is stratified into four difficulty levels: Curriculum-Junior/Senior and Olympiad-Junior/Senior.

\paragraph{GeoAux-Canvas.}
To assess generalization at scale, we adapted samples from \textit{MathCanvas-Bench} \cite{Shi2025MathCanvasIV}. We filtered for construction-heavy problems and applied our annotation pipeline to generate corresponding $(T_{aux}, I_{aux})$ pairs. This subset retains fine-grained subject tags (e.g., Analytic Geometry, Trigonometry), enabling detailed capability analysis.

\paragraph{Quality Control and Standardization.}
We enforced a rigorous three-stage pipeline:
(1) Solvability Verification: Using Gemini-2.5-Pro \cite{Comanici2025Gemini2P} to ensure problem conditions are sufficient for a unique solution;
(2) Symbolic Normalization: Parsing all mathematical expressions into a unified \LaTeX{} format;
(3) Visual Enhancement: We utilized Seedream 4.0 \cite{Chen2025Seedream4T} for super-resolution to enhance image quality, followed by normalization to $512 \times 512$.
Furthermore, to enable deterministic evaluation of open-ended proofs, we reformulate proof problems (e.g., ``Prove $A=B$'') into verifiable computational problems (e.g., ``Find the ratio $A/B$'').

\paragraph{Statistics.}

\begin{figure}[htbp]
    \centering
    \begin{subfigure}[b]{\linewidth}
        \centering
        \includegraphics[width=\linewidth]{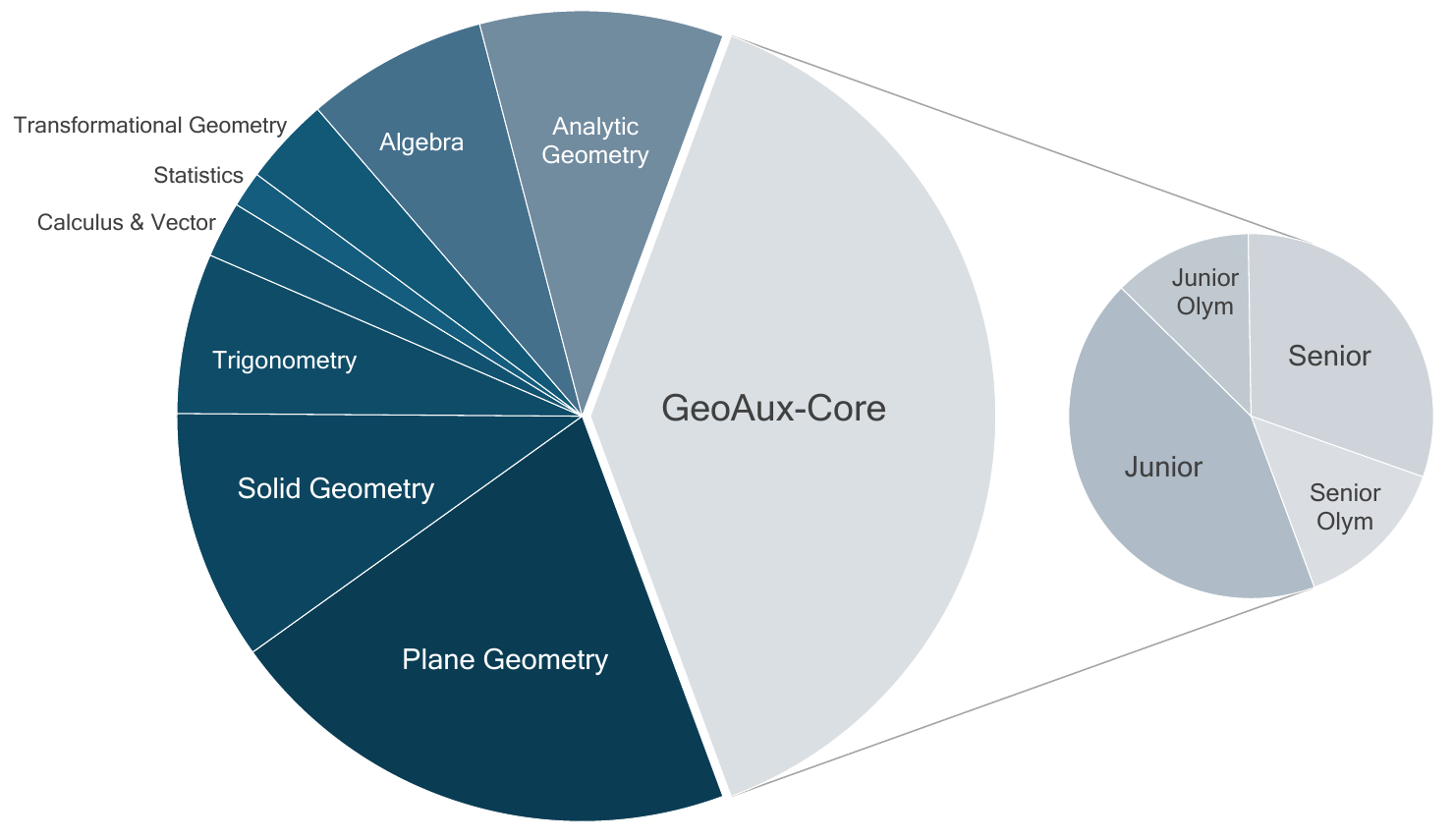} 
        \caption{Subject \& Difficulty Distribution}
    \end{subfigure}
    \vspace{0.2cm}
    \begin{subfigure}[b]{0.395\linewidth} 
        \centering
        \includegraphics[width=\linewidth]{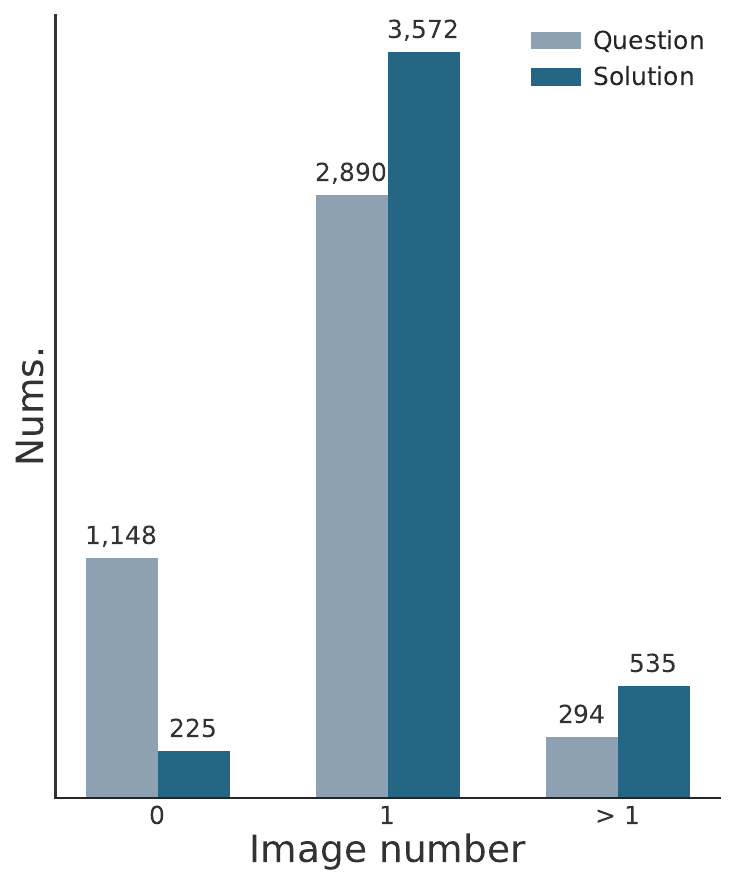}
        \caption{Image Count}
    \end{subfigure}
    \hfill
    \begin{subfigure}[b]{0.5925\linewidth} 
        \centering
        \includegraphics[width=\linewidth]{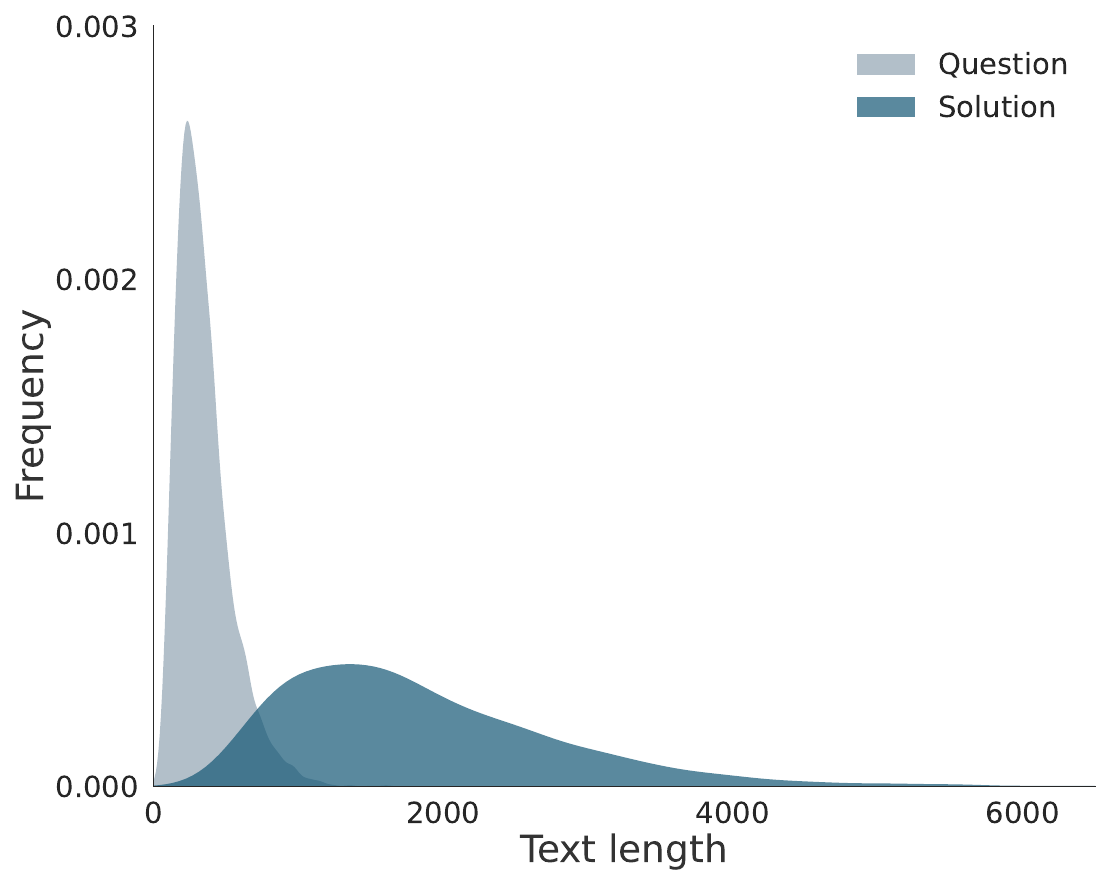}
        \caption{Text Length Distribution}
    \end{subfigure}
    \caption{\textbf{Overview of GeoAux-Bench.}}
    \label{fig:benchmark_stats}
\end{figure}
GeoAux-Bench comprises 4,334 problems (6,523 queries) and 8,470 geometric diagrams, with an extremely high image-to-problem ratio as illustrated in Figure~\ref{fig:benchmark_stats}. Detailed statistics are provided in Appendix~\ref{sec:detailedbench}.

\subsection{Pilot Study: Complementarity of Auxiliary Modalities in Interleaved Geometric Reasoning}
\label{sec:pilot_study}
We conceptualize geometric construction as a critical Interleaved Reasoning Step with dual representations: the \textbf{Auxiliary Construction Instruction} ($T_{aux}$) and the \textbf{Auxiliary Diagram} ($I_{aux}$). 
While recent unified MLLMs (e.g., Chameleon-7B \cite{Team2024ChameleonME}, Bagle-7B-MoT \cite{Deng2025EmergingPI}) attempt end-to-end interleaved generation, their capability to precisely render geometric constraints remains nascent. Consequently, prior works often resort to compromises: they convert diagrams that ought to be presented via the visual modality into single-modal formal languages or drawing code, thereby circumventing the challenge of visual generation. 

This raises a fundamental research question: \textit{Does $T_{aux}$—the textual modality describing the visual diagram—fully encapsulate the latent spatial information contained in the corresponding visual diagram $I_{aux}$, or do these two modalities offer complementary cognitive grounding?}
To disentangle the contributions of the semantic instruction and the corresponding visual diagram, we design a controlled ablation study. Let $\mathcal{Q}$ denote the problem text and $I_{orig}$ the initial diagram. We evaluate performance under four experimental settings:

\begin{itemize}[noitemsep, leftmargin=*]
    \item \textbf{Standard Settings:} $(\mathcal{Q}, I_{orig})$. Baseline reasoning without hints.
    \item \textbf{Textual-Only Settings:} $(\mathcal{Q}, I_{orig}, T_{aux})$. Providing only the auxiliary construction instruction.
    \item \textbf{Visual-Only Settings:} $(\mathcal{Q}, I_{orig}, I_{aux})$. Providing only the auxiliary diagram.
    \item \textbf{Interleaved Settings:} $(\mathcal{Q}, I_{orig}, T_{aux}, I_{aux})$. Simulating an ideal interleaved reasoning step with both modalities.
\end{itemize}

\paragraph{Perceptual Saliency Control (Visual Enhancement).}

\begin{figure}[htp]
  \centering
  \includegraphics[width=0.46\textwidth]{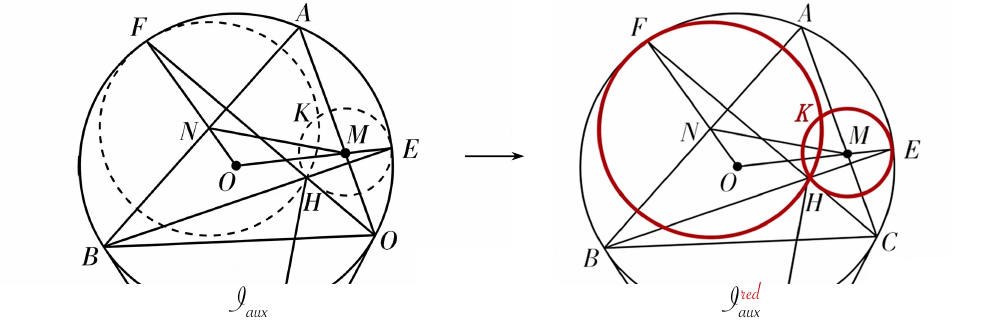}
  \caption{\textbf{Visual Enhancement Protocol.} Bold red lines are used to highlight auxiliary elements ($I_{\text{aux}}^{\text{\textcolor{deepred}{red}}}$).}
  \label{fig:redaux}
\end{figure}

A potential confounder in the \textit{Visual-Only Setting} is visual saliency: standard (typically dashed) auxiliary lines may be too subtle for detection, leading to false negatives from \textit{perceptual misses}.To boost the saliency of these elements and prevent them from being overlooked in complex geometric configurations, we introduce a Visual Enhancement Protocol, as illustrated in Figure~\ref{fig:redaux}. Specifically, we identified 200 hard samples where models succeeded with the \textit{Textual-Only Settings} but failed with the \textit{Visual-Only Settings}; to these, we applied the enhanced annotation (denoted $I_{\text{aux}}^{\text{\textcolor{deepred}{red}}}$) to ensure visibility, yielding two control settings: \textbf{Enhanced Visual-Only} $(\mathcal{Q}, I_{orig}, I_{\text{aux}}^{\text{\textcolor{deepred}{red}}})$ and \textbf{Enhanced Dual-Modal} $(\mathcal{Q}, I_{orig}, T_{aux}, I_{\text{aux}}^{\text{\textcolor{deepred}{red}}})$.

\paragraph{Analysis of Reasoning Dynamics.}

Table~\ref{tab:pilot_results} presents the performance metrics. By analyzing the intra-group trends, we derive three critical insights into the cognitive mechanism of auxiliary construction:

\begin{table}[htbp]
\centering
\definecolor{deepred}{HTML}{8B0000} 

\resizebox{\linewidth}{!}{
\begin{tabular}{@{}l l l r r@{}} 
\toprule
\toprule
\textbf{Setting} & \textbf{Input Modality} & \textbf{Acc\% ($\Delta$)} & \textbf{Tokens} & \textbf{PPL} \\ 
\midrule
\multicolumn{5}{l}{\textit{\textbf{G1: Text-Only Settings}}} \\ 
Standard & $\mathcal{Q},I_{orig}$ & 23.68 \phantom{(+0.00)} & 1336.2 & 1.1250 \\ 
Text-Only & $+T_{aux}$ & 24.24 \textcolor{teal}{(+0.56)} & 1241.3 & \textbf{1.1185} \\ 
\midrule
\multicolumn{5}{l}{\textit{\textbf{G2: Visual-Only Settings}}} \\ 
Visual-Only & $+I_{aux}$ & 23.34 \textcolor{purple}{(-0.34)} & 1258.0 & 1.1340 \\ 
Enhanced Visual & $+I_{\text{aux}}^{\text{\textcolor{deepred}{red}}}$ & 24.44 \textcolor{teal}{(+0.76)} & 1303.6 & 1.1310 \\ 
\midrule
\multicolumn{5}{l}{\textit{\textbf{G3: Interleaved Settings}}} \\ 
Dual-Modal & $+T_{aux}+I_{aux}$ & 24.90 \textcolor{teal}{(+1.22)} & \textbf{1230.7} & 1.1350 \\ 
Enhanced Dual & $+T_{aux}+I_{\text{aux}}^{\text{\textcolor{deepred}{red}}}$ & \textbf{25.31} \textcolor{teal}{(+1.63)} & 1270.5 & 1.1323 \\ 
\bottomrule
\bottomrule
\end{tabular}
}
\caption{Reasoning Performance under Different Modal Input Combinations. Experiments are grouped by context structure for fair intra-group comparisons.}
\label{tab:pilot_results}
\end{table}

\noindent(1) \textbf{Irreplaceability and Complementarity of Modalities.}
While the \textit{Text-Only Setting} outperforms the baseline, it still falls 1.07\% short of the \textit{Interleaved Setting}. This directly answers our research question: Textual instructions ($T_{\text{aux}}$) cannot fully replace visual diagrams ($I_{\text{aux}}$). 
The two modalities provide complementary cognitive grounding: text defines operational instructions to guide focus on target elements and resolve semantic ambiguity, while diagrams deliver spatial relationships to clarify geometric connections and reduce uncertainty. Optimal reasoning performance arises only when the model ``sees'' the spatial realization of its intent.

\noindent(2) \textbf{Lower PPL Correlates with Higher Accuracy Across Modal Settings.}
Across all groups, we observe a strong positive correlation between reduced Perplexity and improved reasoning accuracy under the same modal configuration—notably, with no significant correlation between accuracy and generated token length. This aligns with human geometric reasoning intuition: just as a well-constructed auxiliary line triggers an "epiphany" that simplifies a complex problem, valid auxiliary information lowers the model’s predictive entropy. A lower PPL may signals a clearer, less ambiguous solution path, which empirically boosts the likelihood of correct reasoning—an insight we leverage in our reward shaping design.

\noindent(3) \textbf{Visual Saliency Matters.}
Comparing \textit{Visual-Only} with \textit{Enhanced Visual} (and \textit{Interleaved} analogously), enhancing visual saliency consistently reduces PPL and improves Accuracy (up to +1.10\%). Notably, this performance gain stems from modifying only 200 specific samples. This sensitivity indicates the model’s performance is highly reliant on the perceptual clarity of auxiliary elements. We hypothesize visual feature enhancement enables the model’s visual system to more reliably detect and anchor geometric configuration changes. This confirms visual perceptibility is a strict prerequisite for geometric reasoning. For qualitative analysis of attentional patterns, see Appendix~\ref{sec:appendix_mechanism}.

\section{Method}
\label{sec:method}
\begin{figure*}[htbp] 
  \centering
  \includegraphics[width=0.95\linewidth]{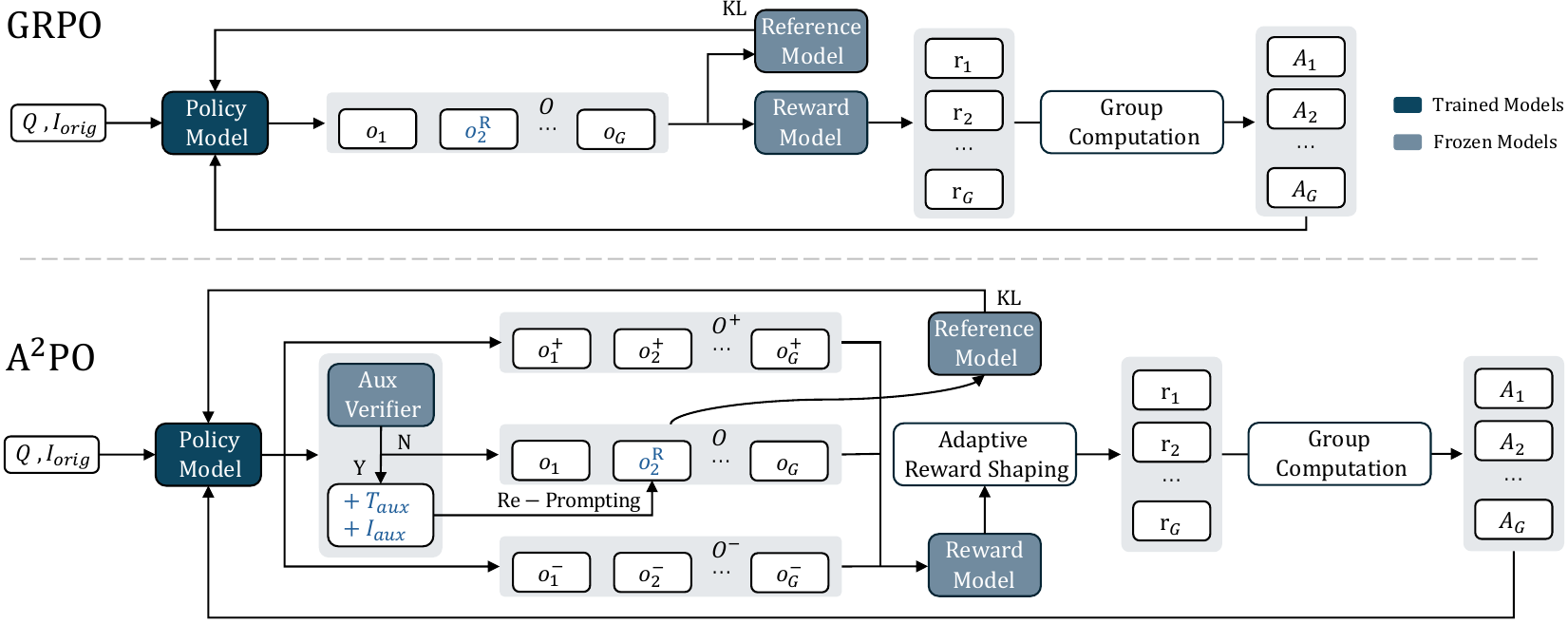} 
  \caption{The framework of Action Applicability Policy Optimization (A\textsuperscript{2}PO). The upper panel shows standard GRPO, while the lower panel illustrates our tri-partition sampling and adaptive reward shaping mechanism.} 
  \label{fig:A2PO} 
\end{figure*}

\subsection{Preliminaries: GRPO}
\label{sec:grpo_prelim}

We adopt Group Relative Policy Optimization (GRPO) as our optimization backbone. GRPO eliminates the separate value function by estimating the baseline from the group average.

Formally, given a question $q$, the policy generates a group of outputs $\{o_i\}_{i=1}^G$. The advantage $A_i$ is estimated by normalizing group rewards:

\begin{small}
\begin{equation}
\small
A_i = \frac{R_i - \text{mean}(\{R_j\}_{j=1}^G)}{\text{std}(\{R_j\}_{j=1}^G) + \epsilon},
\label{eq:grpo_advantage}
\end{equation}
\end{small}

\noindent where $\epsilon$ is a small constant. GRPO optimizes the surrogate objective, averaged over tokens:
{

\small

\begin{equation}
\begin{split}
\mathcal{J}_{GRPO}(\theta) = \mathbb{E}_{q, \mathbf{o}} \Bigg[ \frac{1}{G} \sum_{i=1}^G \frac{1}{|o_i|} \sum_{t=1}^{|o_i|} \bigg( \min \Big( r_{i,t} A_i, \\
\text{clip}(r_{i,t}, 1\!-\!\varepsilon, 1\!+\!\varepsilon) A_i \Big) - \beta \mathbb{D}_{KL}(\pi_\theta || \pi_{\text{ref}}) \bigg) \Bigg],
\end{split}
\label{eq:grpo_objective}
\end{equation}
}
\noindent where $r_{i,t} = \frac{\pi_\theta(o_{i,t} | q, o_{i,<t})}{\pi_{\theta_{old}}(o_{i,t} | q, o_{i,<t})}$ is the probability ratio, and $\mathbf{o}=\{o_i\}_{i=1}^G$ denotes the sampled group.

\subsection{Overview of A\textsuperscript{2}PO}
Building upon GRPO, we propose \textbf{Action Applicability Policy Optimization (A\textsuperscript{2}PO)}. While standard GRPO optimizes solely for outcome correctness, geometric reasoning requires mastering the strategic applicability of auxiliary constructions—specifically, discerning whether a visual modification is beneficial for the specific problem configuration.
To explicitly model this applicability, A\textsuperscript{2}PO introduces a \textbf{Tri-Partition Sampling} mechanism. As illustrated in Figure~\ref{fig:A2PO}, instead of sampling from a single policy distribution, we partition the rollout trajectories into three distinct subsets. This structure enables the construction of counterfactual baselines—comparing scenarios where auxiliary lines are enforced versus prohibited—to derive an \textbf{Adaptive Reward} that guides the model on \textit{when} to construct (Timing) and \textit{how} to construct efficiently (Quality).

\subsection{Tri-Partition Sampling with Visual Re-prompting}
\label{sec:sampling}

To quantify the marginal utility of auxiliary constructions, we sample trajectories via three distinct conditioning protocols, aggregating them into a rollout set $\mathcal{G} = \{O^+, O^-, O\}$. For a given query $q$ and initial diagram $I_{\text{orig}}$, we sample $N$ trajectories for each subset. Prompts for each protocol are provided in Appendix~\ref {app:prompts}:

\begin{itemize}[noitemsep,leftmargin=*]
    \item \textbf{Mandatory Subset ($O^+$):} We employ Prefix Forcing to enforce geometric intervention. Given that the model is fine-tuned during the supervised warm-up to encapsulate auxiliary commands in \texttt{<aux>} tags, we pre-populate the generation prefix with \texttt{<aux>}. This strictly forces a valid auxiliary construction step while retaining the standard prompt context.
    \item \textbf{Prohibited Subset ($O^-$):} We impose a Hard Constraint to disable visual updates. In addition to appending a negative constraint to the system prompt, we explicitly mask the logits of the \texttt{<aux>} and \texttt{</aux>} tokens during decoding, guaranteeing a reasoning trajectory devoid of auxiliary constructions and forcing the model to rely solely on the initial visual context.
    \item \textbf{Natural Subset ($O$):} The model samples autonomously using the same standard prompt as $G^+$, without any intervention. This subset represents the target policy to be optimized.
\end{itemize}

\paragraph{Visual Re-prompting.}
To simulate interleaved visual reasoning despite current rendering limitations, we employ a retrieval-based injection strategy within the Natural Subset. Upon detecting a completed auxiliary command, an Aux Verifier evaluates its semantic equivalence to the ground truth. If and only if the construction is equivalent to the ground truth, we trigger re-prompting: the generation is paused, and the model is queried again with an augmented context that appends a structured ``Hint'' containing the ground-truth instruction ($T_{\text{aux}}$) and the corresponding auxiliary diagram ($I_{\text{aux}}$). This provides high-fidelity visual feedback contingent on correct reasoning. Prompt transformations are detailed in Appendix~\ref{app:prompts}.

\subsection{Adaptive Reward Shaping}
\label{sec:rewards}

We design a composite reward function $R(o)$ specifically for the natural subset $O$:
{\small
\begin{equation}
R(o) = w_{1} r_{acc} + w_{2} r_{fmt} + w_{3} r_{time} + w_{4} r_{qual}
\end{equation}
}
where $w_{(\cdot)}$ are weighting coefficients. While $r_{acc}$ and $r_{fmt}$ align with standard GRPO protocols, we introduce $r_{time}$ and $r_{qual}$ to optimize auxiliary construction efficacy:

\subsubsection{Timing Reward ($r_{time}$)}
This component trains the model to discern the strategic necessity of auxiliary constructions. Let $\mathbb{I}_{aux}(o)$ be the auxiliary indicator and $\Delta = \mathbb{E}_{O^+}[r_{acc}] - \mathbb{E}_{O^-}[r_{acc}]$ be the utility gap. We formulate $r_{time}(o)$ with a significance margin $\tau$:
{
\small
\begin{equation}
r_{time}(o) = \mathbb{I}_{aux}(o) \cdot
\begin{cases} 
    1 & \text{if } \Delta > \tau \\
    -1 & \text{if } \Delta < -\tau \\
    0 & \text{otherwise}
\end{cases}
\label{eq:timing_reward}
\end{equation}
}

\noindent This restricts auxiliary construction to scenarios yielding a net performance gain.

\subsubsection{Quality Reward ($r_{qual}$)}
Building on our pilot findings (Sec.~\ref{sec:pilot_study}), we posit that effective auxiliary constructions function as \textbf{entropy reducers}, mirroring the ``cognitive epiphany'' that manifests as reduced reasoning PPL.

We establish a perplexity baseline $\bar{\mathcal{P}} = \mathbb{E}_{O^+}[\text{PPL}]$ derived from the mandatory subset. The quality reward grants a bonus solely to valid, low-entropy auxiliary usage:

{ \small
\begin{equation}
r_{qual}(o) = \mathbb{I}_{aux}(o) \cdot r_{acc}(o) \cdot \mathbb{I}\big(\text{PPL}(o) < \bar{\mathcal{P}} + \delta\big)
\label{eq:quality_reward}
\end{equation}
}
\noindent where $\delta$ compensates for the PPL overhead induced by visual re-prompting. This reward explicitly favors confident reasoning paths that simplify the solution space. Detailed hyperparameter configurations are provided in Appendix~\ref{app:hyperparams}.
\section{Experiments}
We split our experiments into two distinct parts to rigorously validate our contributions: (1) conducting a systematic evaluation of popular MLLMs on GeoAux-Bench to establish a difficulty baseline; (2) focusing on our proposed A\textsuperscript{2}PO framework, with comparisons against strong RL baselines across multiple benchmarks to verify the efficacy of our adaptive reward shaping.

\subsection{Benchmark Results}
\label{sec:benchmark_results}

We conduct a comprehensive evaluation on GeoAux-Bench involving various popular MLLMs, including proprietary SOTA models \cite{Comanici2025Gemini2P,openai_gpt5_2025,Hurst2024GPT4oSC,anthropic_claude_opus4_1_2025,anthropic_claude_sonnet4_5_2025,bytedance_seed1_6_2025}, open-weights baselines \cite{Bai2025Qwen3VLTR,Bai2025Qwen25VLTR,Zhu2025InternVL3EA,Wang2025InternVL35AO,Hong2025GLM45VAG}, and native unified models capable of interleaved image-text generation \cite{Shi2025MathCanvasIV,Deng2025EmergingPI,Li2025ZebraCoTAD}.
All models are evaluated under a unified setting (see Appendix \ref{app:hyperparams} for hyperparameters). The performance on GeoAux-Bench-Core is reported in Table \ref{tab:benchmark_results}. Results on the GeoAux-Bench-Canvas subset are provided in Appendix \ref{sec:detailedbench}.

\begin{table}[t]
\centering
\setlength{\tabcolsep}{3pt} 

\resizebox{\linewidth}{!}{
    \begin{tabular}{l c c c c c c}
    \toprule
    \toprule
    \multirow{2}{*}{\textbf{Model}} & \multirow{2}{*}{\textbf{Think}} & \multicolumn{2}{c}{\textbf{Curriculum}} & \multicolumn{2}{c}{\textbf{Olympiad}} & \multirow{2}{*}{\textbf{Total}} \\ 
    \cmidrule(lr){3-4} \cmidrule(lr){5-6} 
    
     & & \textbf{Senior} & \textbf{Junior} & \textbf{Senior} & \textbf{Junior} & \\ 
    \midrule
    
    \multicolumn{7}{c}{\textit{Closed-source MLMMs}} \\
    \midrule
    Gemini-2.5-Pro & \ding{51} & \cellcolor{bestclosed}\textbf{83.91} & \cellcolor{bestclosed}\textbf{84.75} & \cellcolor{bestclosed}\textbf{82.28} & 88.24 & \cellcolor{bestclosed}\textbf{83.16} \\
    Gemini-2.5-Flash & \ding{51} & 80.10 & 81.68 & 64.56 & 82.13 & 79.53 \\
    GPT-5 & \ding{51} & 71.48 & 84.02 & 79.32 & \cellcolor{bestclosed}\textbf{88.41} & 80.62 \\
    GPT-4o & \ding{55} & 19.57 & 24.29 & 57.38 & 42.51 & 28.13 \\
    Claude-Opus-4.1 & \ding{51} & 60.36 & 52.70 & 58.23 & 58.45 & 55.82 \\
    Claude-Sonnet-4.5 & \ding{51} & 59.20 & 52.00 & 55.70 & 60.39 & 55.03 \\
    Doubao-seed-1.6 & \ding{51} & 62.02 & 61.42 & 72.57 & 86.47 & 65.00 \\
    
    \midrule
    \multicolumn{7}{c}{\textit{Open-source MLLMs}} \\
    \midrule
    Qwen3-VL-235B-Ins. & \ding{55} & 61.36 & \cellcolor{bestopen}\textbf{77.80} & 78.90 & 91.79 & \cellcolor{bestopen}\textbf{74.85} \\
    Qwen3-VL-235B-Thk. & \ding{51} & \cellcolor{bestopen}\textbf{64.68} & 50.36 & \cellcolor{bestopen}\textbf{89.45} & \cellcolor{bestopen}\textbf{95.17} & 62.25 \\
    Qwen3-VL-30B-Ins. & \ding{55} & 61.36 & 70.38 & 75.53 & 89.37 & 70.25 \\
    Qwen3-VL-30B-Thk. & \ding{51} & 55.72 & 49.64 & 83.97 & 93.24 & 58.75 \\
    Qwen2.5-VL-Ins. & \ding{55} & 23.05 & 28.81 & 65.82 & 52.17 & 33.25 \\
    InternVL3.5-8B & \ding{55} & 38.31 & 26.63 & 54.85 & 66.67 & 36.26 \\
    InternVL3-8B & \ding{55} & 25.87 & 30.02 & 67.93 & 55.56 & 35.17 \\
    GLM-4.1V-Thk. & \ding{51} & 31.67 & 23.57 & 56.12 & 53.14 & 31.76 \\
    GLM-4.5V & \ding{55} & 45.27 & 24.62 & 72.15 & 82.61 & 40.24 \\
    MiMo-VL-7B-SFT & \ding{55} & 50.41 & 42.13 & 71.73 & 80.68 & 50.87 \\
    MiMo-VL-7B-RL & \ding{51} & 50.75 & 41.65 & 71.73 & 80.68 & 50.70 \\
     \midrule
    \multicolumn{7}{c}{\textit{United MLLMs}} \\
    \midrule
    
    BAGEL-7B-MoT & \ding{55} & 7.30 & 7.34 & 34.60 & 38.16 & 12.95 \\
    BAGEL-Zebra-CoT  & \ding{55} & 6.80 & 6.94 & 31.65 & 35.27 & 12.03 \\
    MathCanvas-7B & \ding{55} & 18.41 & 18.89 & 46.84 & 51.69 & 24.63 \\
    \bottomrule
    \bottomrule
    \end{tabular}
}
\caption{Comparison on \textbf{GeoAux-Bench-Core}. Best \colorbox{bestclosed}{\textbf{closed}} and \colorbox{bestopen}{\textbf{open}} scores are highlighted. ``Ins.'' and ``Thk.'' denote Instruct and Thinking models.}
\label{tab:benchmark_results}
\end{table}

\textbf{Analysis.}
The results highlight three critical observations:

\noindent(1) \textbf{Performance Gap \& Difficulty.} 
A substantial performance chasm exists between top-tier proprietary models (e.g., Gemini-2.5-Pro at 83.16\%) and typical open-weights baselines (e.g., InternVL3.5-8B at 36.26\%). Furthermore, the benchmark proves exceptionally difficult for Native Unified MLLMs (e.g., BAGEL < 13\%), which perform the worst among all paradigms. Despite being intuitively aligned with human ``thinking-while-drawing,'' qualitative analysis (see Appendix \ref{app:case1}) reveals that these models fail to execute precise geometric edits, leading to severe visual hallucinations that derail subsequent reasoning.

\noindent(2) \textbf{The Analytic Shortcut.} 
Models frequently outperform on Senior geometry compared to Junior geometry. Case studies in Appendix \ref{app:case2} reveal that MLLMs exhibit a strong preference for \textit{Analytic Geometry} (e.g., establishing coordinate systems) to bypass visual intuition. While this algebraic conversion works for high school problems, it is often inefficient or inapplicable to the pure geometric logic required in Junior tasks, explaining the performance inversion.

\noindent(3) \textbf{Signs of Memorization.} 
Two anomalies point to potential data contamination rather than robust reasoning. First, models like Qwen3-VL-Thk achieve remarkable scores on the challenging Senior Olympiad split (89.45\%) yet struggle significantly on the simpler Junior Curriculum set (50.36\%). Second, reasoning-enhanced ``Thinking'' models often underperform their instruction-tuned counterparts. We attribute this to the \textit{finite and public nature} of Olympiad problems. Unlike the vast space of curriculum exercises, high-profile competition questions are limited in number and widely circulated, making them highly susceptible to inclusion in pre-training corpora. This suggests that such performance spikes likely stem from memorizing specific problem instances rather than generalized geometric reasoning.

\subsection{Performance of A\textsuperscript{2}PO}
We evaluate our proposed A\textsuperscript{2}PO against SFT and strong RL baselines (GRPO \cite{Shao2024DeepSeekMathPT}, ToRL \cite{Li2025ToRLST}, GeometryZero \cite{Wang2025GeometryZeroIG}). Experiments use Qwen2.5-VL across three datasets: GeoAux-Bench, external benchmarks Geomverse \cite{Kazemi2023GeomVerseAS} and Geometry3k \cite{Lu2021InterGPSIG}. Results in Table \ref{tab:main_experiment}; training/inference configurations in Appendix \ref{app:hyperparams}.

\begin{table}[htbp]
\centering
\small
\resizebox{\linewidth}{!}{
    \begin{tabular}{l c c c c | c} 
    \toprule
    \toprule
    \multirow{2}{*}{\textbf{Method}} & \textbf{GeoAux} & \textbf{Geomverse} & \textbf{Geometry3k} & \textbf{Avg.} & \textbf{GeoAux}  \\
     & Acc\% & Acc\% & Acc\% & Acc\% & PPL $\downarrow$ \\
    \midrule
    \multicolumn{6}{c}{\textit{Qwen2.5-VL-3B-Instruct}} \\
    \midrule
    SFT & 23.09 & 56.20 & 39.40 & 39.56 & \cellcolor{bestopen}\textbf{1.1389} \\
    GRPO & \textbf{31.22} & \cellcolor{bestopen}\textbf{59.10} & 50.72 & \textbf{47.01} & 1.1550 \\
    ToRL & 28.68 & 58.40 & 47.06 & 44.71 & 1.1558 \\
    GeometryZero & 29.33 & 57.00 & \textbf{52.72} & 46.35 & 1.1535 \\
    \textbf{A\textsuperscript{2}PO (Ours)} & \cellcolor{bestopen}\textbf{33.20} & 58.40 & \cellcolor{bestopen}\textbf{53.05} & \cellcolor{bestopen}\textbf{48.22} & \textbf{1.1534} \\
    \midrule
    \multicolumn{6}{c}{\textit{Qwen2.5-VL-7B-Instruct}} \\
    \midrule
    SFT & 37.30 & 63.60 & 46.17 & 49.02 & \cellcolor{bestopen}\textbf{1.0857} \\
    GRPO & 39.28 & 67.40 & \cellcolor{bestopen}\textbf{55.49} & 54.06 & 1.0887 \\
    ToRL & 39.77 & 65.50 & 53.50 & 52.92 & 1.0941 \\
    GeometryZero & \textbf{40.18} & \textbf{68.30} & \textbf{53.72} & \textbf{54.07} & 1.0945 \\
    \textbf{A\textsuperscript{2}PO (Ours)} & \cellcolor{bestopen}\textbf{42.97} & \cellcolor{bestopen}\textbf{70.70} & 53.61 & \cellcolor{bestopen}\textbf{55.76} & \textbf{1.0869} \\
    \bottomrule
    \bottomrule
    \end{tabular}
}
\caption{Main results on GeoAux and external benchmarks. \colorbox{bestopen}{\textbf{Best}} and \textbf{Second Best} results are highlighted.}
\label{tab:main_experiment}
\end{table}

\textbf{Analysis.}
The comparative results demonstrate the superiority of our policy optimization strategy:

\noindent(1) \textbf{Consistent Improvements.} 
A\textsuperscript{2}PO consistently outperforms all baselines across model scales. Notably, on the 7B scale, A\textsuperscript{2}PO achieves an average accuracy of 55.76\%, surpassing the strong GeometryZero baseline (54.07\%) and standard GRPO (54.06\%). This gain is most pronounced on GeoAux (+2.79\% over GeometryZero), confirming that our reward shaping is particularly effective for geometric problems that heavily rely on auxiliary construction for their solution.

\noindent(2) \textbf{Reasoning Certainty \& Efficiency.} 
A critical observation lies in Perplexity: strong baselines like GeometryZero and ToRL improve accuracy but see elevated PPL compared to SFT (e.g., up to +0.0169),suggesting performance gains may come at the cost of uncertain or convoluted reasoning. 
In contrast, A\textsuperscript{2}PO achieves the highest accuracy while maintaining a PPL nearly matching SFT. 
This validates Section \ref{sec:pilot_study}: auxiliary constructions act as ``cognitive scaffolds'' reducing uncertainty. The Quality Reward guides A\textsuperscript{2}PO toward \textit{elegant} simplifications—akin to a geometric ``epiphany''—over convoluted computation.

\subsection{Ablation Study}
Table \ref{tab:ablation_components} presents a component-wise ablation on Qwen2.5-VL-7B-Instruct to validate the efficacy of each component.

\begin{table}[htbp]
\centering
\resizebox{\linewidth}{!}{
    \begin{tabular}{l l c c c c c }
    \toprule
    \toprule
   \textbf{Base Model} & \textbf{Method} & \textbf{LR} & \textbf{TR} & \textbf{QR} & \textbf{Vis} & \textbf{Acc\%} \\
    
    \midrule
    
    \multirow{5}{*}{\shortstack[l]{\textit{Qwen2.5-7B-Instruct}}} 
    & GRPO & \ding{51} & \ding{55} & \ding{55} & \ding{55} & 39.28 \\
    
    & GRPO(w/o LR)& \ding{55} & \ding{55} & \ding{55} & \ding{55} & 39.52 \\
    
    & A\textsuperscript{2}PO (w/o TR, w/o Vis) & \ding{55} & \ding{51} & \ding{55} & \ding{55} & 40.18 \\
    & A\textsuperscript{2}PO (w/o Vis) & \ding{55} & \ding{51} & \ding{51} & \ding{55} & 41.17 \\
    
    \rowcolor{bestopen}& \textbf{A\textsuperscript{2}PO} & \ding{55} & \ding{51} & \ding{51} & \ding{51} & \textbf{42.97} \\
    
    \bottomrule
    \bottomrule
    \end{tabular}
}
\caption{Ablation study of A\textsuperscript{2}PO components, including Length Reward (LR), Timing Reward (TR), Quality Reward (QR), and Visual Re-prompting (Vis). Accuracy results are based on GeoAux-Bench.}
\label{tab:ablation_components}
\end{table}

\textbf{Analysis.}
We analyze the incremental impact of our design choices:

\noindent(1) \textbf{Removing Length Bias (LR).} 
Removing the standard Length Reward yields a performance gain (39.28\% $\rightarrow$ 39.52\%). Unlike open-ended generation where length often correlates with detail, mathematical proofs prioritize \textit{conciseness}. We observed that generic length incentives inadvertently encouraged verbose behaviors (e.g., repetitive reasoning). Removing this constraint allows the model to focus on the logical precision of the solution rather than artificially extending the trajectory.

\noindent(2) \textbf{Efficacy of Reward Shaping (TR \& QR).} 
Incorporating the Timing Reward (TR) improves accuracy to 40.18\%, explicitly surpassing the ToRL baseline (39.77\%). While ToRL indiscriminately rewards \textit{any valid auxiliary construction}, TR teaches the model the strategic distinction of \textit{necessity}—rewarding the action only when it is strictly beneficial compared to a non-auxiliary path. 
Further adding the Quality Reward (QR) pushes performance to 41.17\%. This confirms that guiding the model towards lower-perplexity (i.e., more confident) constructions effectively filters out \textit{valid-but-redundant} auxiliary lines—constructions that are syntactically correct but offer no strategic value to the solution.

\noindent(3) \textbf{Visual Synergy (Vis).} 
The most significant jump occurs with \textbf{Visual Re-prompting} (+1.80\%), achieving the peak accuracy of 42.97\%. This empirical evidence strongly supports our core contribution: textual descriptions of auxiliary lines alone are insufficient. The model achieves optimal reasoning only when the textual construction is explicitly rendered and injected back as an updated visual context. This mechanism successfully simulates the cognitive advantage of ``thinking with images'' within a re-prompting framework.
\section{Conclusion}
In this work, we advance geometric problem solving by shifting from static perception to \textbf{active Visual-Text Interleaved reasoning}.
First, we introduce \textbf{GeoAux-Bench}, the first benchmark to rigorously align textual construction instructions with corresponding visual updates, providing a precise testbed for multimodal interaction.
Second, our empirical analysis reveals that visual aids function as essential \textbf{entropy reducers}: interleaved visual feedback significantly lowers reasoning uncertainty compared to single-modality inputs, validating the cognitive necessity of ``thinking with images.''
Capitalizing on this, we propose \textbf{Action Applicability Policy Optimization (A\textsuperscript{2}PO)}. By employing \textbf{Adaptive Reward Shaping} to strictly regulate the \textit{timing} and \textit{quality} of visual interventions, A\textsuperscript{2}PO enables models to master the strategic deployment of auxiliary lines, achieving state-of-the-art performance.
Our findings demonstrate that equipping models with the agency to actively modify their visual context is a pivotal step toward autonomous, physically grounded geometric reasoning.
\section*{Limitations}
The primary limitation of this work lies in the implementation of the visual update mechanism. While our A\textsuperscript{2}PO framework establishes the cognitive benefits of interleaved reasoning, the current execution relies on a retrieval-based injection strategy—utilizing ground-truth auxiliary diagrams—rather than native model-generated visuals.
This design choice is necessitated by the capabilities of current Unified MLLMs. Despite recent advancements, existing models still lack the fine-grained visual actuation capability required for precise geometric editing. As observed in our pilot experiments, even atomic operations (e.g., ``\textit{Connect points A and B}'') frequently result in structural hallucinations or pixel-level inaccuracies that can propagate errors into the reasoning chain. Furthermore, the high inference latency associated with autoregressive image generation currently hinders efficient exploration during reinforcement learning.
Consequently, our work simulates the \textit{strategic decision-making} of visual construction rather than the \textit{physical execution}. We posit that realizing a fully autonomous loop—where a model generates, perceives, and refines its own diagrams—requires future advancements in multimodal pre-training. Specifically, it demands a tighter alignment between high-level geometric conceptual understanding and low-level pixel editing skills. Only upon this foundation can true dynamic sampling and reinforcement learning for generative visual-text reasoning be achieved.

\bibliography{custom}

\appendix
\section{Qualitative Analysis of Attentional Patterns}
\label{sec:appendix_mechanism}

To complement the quantitative results in Section \ref{sec:pilot_study}, we present the attentional patterns of Qwen2.5-VL-7B-Instruct for two representative examples before and after visual saliency enhancement (bold red auxiliary lines). No significant attentional shift toward the red-highlighted auxiliary elements is observed across cases. However, in both examples, higher attention scores are seen for the alphabetic labels of points involved in auxiliary line construction (e.g., G, M).
\begin{figure}[h]
    \centering
    \begin{subfigure}[b]{0.47\linewidth}
        \centering
        \includegraphics[width=\linewidth]{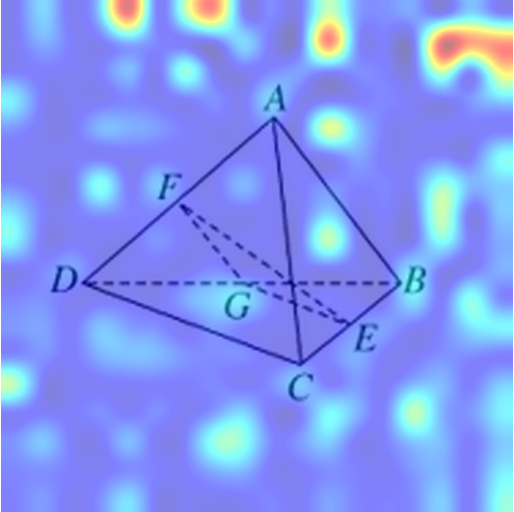}
        \caption{Example 1: Before enhancement}
        \label{fig:ex1_before}
    \end{subfigure}
    \hfill 
    \begin{subfigure}[b]{0.47\linewidth}
        \centering
        \includegraphics[width=\linewidth]{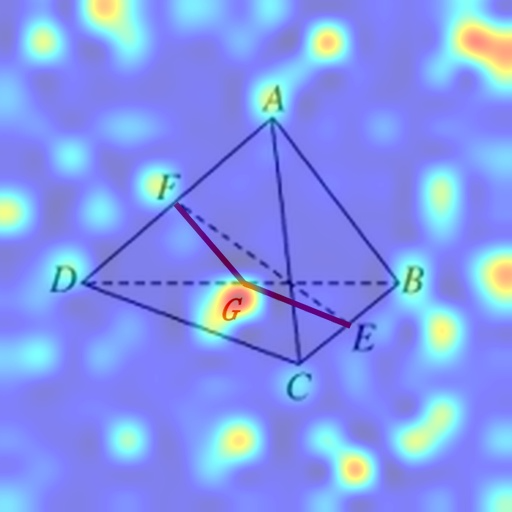}
        \caption{Example 1: After enhancement}
        \label{fig:ex1_after}
    \end{subfigure}

    \vspace{6pt} 

    \begin{subfigure}[b]{0.47\linewidth}
        \centering
        \includegraphics[width=\linewidth]{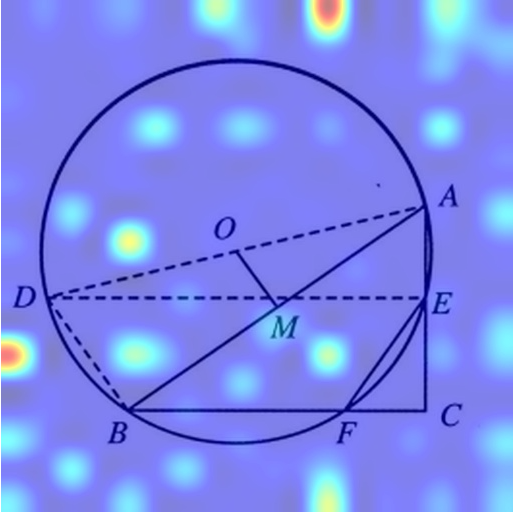}
        \caption{Example 2: Before enhancement}
        \label{fig:ex2_before}
    \end{subfigure}
    \hfill
    \begin{subfigure}[b]{0.47\linewidth}
        \centering
        \includegraphics[width=\linewidth]{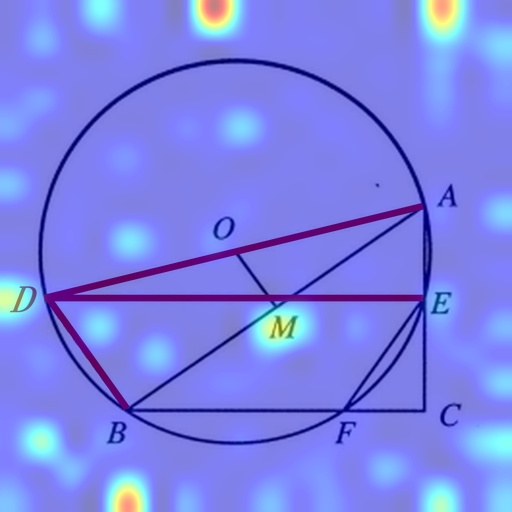}
        \caption{Example 2: After enhancement}
        \label{fig:ex2_after}
    \end{subfigure}

\caption{Attentional heatmaps of Qwen2.5-VL-7B-Instruct. Warmer colors indicate higher attention levels.}
    \label{fig:attention_comparison}
\end{figure}

These observations indicate that visual enhancement may modulate attention via textual point labels rather than the auxiliary lines themselves. However, the underlying mechanism through which a minimal set of visually enhanced samples yields such substantial accuracy gains remains to be further explored.

\section{Detailed Statistics of GeoAux-Bench}
\label{sec:detailedbench}
\begin{table*}[htbp]
\centering
\small
\setlength{\tabcolsep}{5pt} 
\begin{tabular}{@{}l rrrr r r r@{}} 
\toprule
\toprule
\multirow{3}{*}{\textbf{Statistic}} & \multicolumn{5}{c}{\textbf{GeoAux-Core}} & \multirow{3}{*}{\textbf{GeoAux-Canvas}} & \multirow{3}{*}{\textbf{Total}} \\ 
\cmidrule(lr){2-6} 

 & \multicolumn{2}{c}{\textbf{Curriculum}} & \multicolumn{2}{c}{\textbf{Olympiad}} & \multirow{2}{*}{\textbf{Subtotal}} & & \\ 
\cmidrule(lr){2-3} \cmidrule(lr){4-5} 

 & Senior & Junior & Senior & Junior & & & \\ 
\midrule

Problems & 516 & 722 & 234 & 207 & 1,679 & 2,655 & \textbf{4,334} \\ 

Questions & 647 & 1,288 & 237 & 207 & 2,379 & 4,144 & \textbf{6,523} \\ 

Problem Diagrams & 422 & 1,097 & 207 & 206 & 1,932 & 1,693 & \textbf{3,625} \\ 

Solution Diagrams & 505 & 892 & 205 & 199 & 1,801 & 3,044 & \textbf{4,845} \\ 

\bottomrule
\bottomrule
\end{tabular}
\caption{Detailed composition of \textbf{GeoAux-Bench}. The table breaks down the dataset by subset source, difficulty level, and data modality counts.}
\label{tab:bench_stats}
\end{table*}
Table~\ref{tab:bench_stats} presents the comprehensive statistical breakdown of the GeoAux-Bench dataset. 
The benchmark is composed of two primary subsets:
\begin{enumerate}
    \item \textbf{GeoAux-Core:} This expert-curated subset (1,679 problems) is stratified by difficulty source, distinguishing between standard \textbf{Curriculum} problems and \textbf{Olympiad}-level competitions across both Senior and Junior grades.
    \item \textbf{GeoAux-Canvas:} This subset includes 2,655 scale-augmented samples adapted from MathCanvas-Bench to enhance domain diversity.
\end{enumerate}
We report metrics across four dimensions: the number of distinct \textit{Problems}, total \textit{Questions} (including sub-questions), original \textit{Problem Diagrams}, and the generated \textit{Solution Diagrams} (visual auxiliary lines). 

\section{Results on GeoAux-Bench-Canvas}
\label{sec:appendix_canvas}

In this section, we present the supplementary evaluation results on the \textbf{GeoAux-Bench-Canvas} subset. The performance metrics for the evaluated models are directly sourced from \citet{Shi2025MathCanvasIV}. Detailed breakdowns across different geometric sub-domains are provided in Table \ref{tab:canvas_results}.

\begin{table*}[t]
\centering
\small
\setlength{\tabcolsep}{2.5pt} 
\resizebox{\linewidth}{!}{
    \begin{tabular}{l c c c c c c c c c c}
    \toprule
    \toprule
    \multirow{2}{*}{\textbf{Model}} & \multirow{2}{*}{\textbf{Think}} & \textbf{Alg.} & \textbf{Ana.} & \textbf{Calc.} & \textbf{Plane} & \textbf{Solid} & \textbf{Stats.} & \textbf{Trans.} & \textbf{Trig.} & \multirow{2}{*}{\textbf{Total}} \\
    & & (Algebra) & (Geom.) & (Vector) & (Geom.) & (Geom.) & & (Geom.) & & \\
    \midrule

    \multicolumn{11}{c}{\textit{Closed-source MLLMs}} \\
    \midrule
    Gemini-2.5-Pro & \ding{51} & 68.0 & \cellcolor{bestclosed}\textbf{59.2} & 60.2 & \cellcolor{bestclosed}\textbf{54.8} & \cellcolor{bestclosed}\textbf{48.7} & 64.5 & \cellcolor{bestclosed}\textbf{58.5} & \cellcolor{bestclosed}\textbf{69.9} & \cellcolor{bestclosed}\textbf{47.9} \\
    Gemini-2.5-Flash & \ding{51} & 63.2 & 56.5 & 54.6 & 40.7 & 40.7 & 61.1 & 46.8 & 64.6 & 39.3 \\
    Gemini-2.0-Flash & \ding{55} & 39.1 & 32.6 & 38.9 & 31.1 & 25.6 & 51.4 & 28.1 & 38.0 & 21.2 \\
    GPT-4.1 & \ding{55} & 40.4 & 30.7 & 37.1 & 24.1 & 25.1 & 54.0 & 21.5 & 42.5 & 19.0 \\
    GPT-4.1-mini & \ding{55} & 35.7 & 30.5 & 36.5 & 22.0 & 22.4 & 24.8 & 19.7 & 30.3 & 14.6 \\
    GPT-4o & \ding{55} & 21.6 & 17.7 & 21.8 & 19.5 & 18.6 & 17.4 & 13.2 & 23.0 & 9.9 \\
    GPT-5 & \ding{51} & \cellcolor{bestclosed}\textbf{68.7} & 55.5 & \cellcolor{bestclosed}\textbf{64.2} & 45.6 & 36.1 & 64.5 & 42.7 & 66.5 & 43.5 \\
    Claude-Sonnet-4 & \ding{51} & 44.8 & 38.9 & 49.3 & 33.8 & 33.0 & 46.9 & 30.3 & 47.6 & 25.0 \\
    Seed-1.6-Thinking & \ding{51} & 67.7 & 57.5 & 55.9 & 52.2 & 45.0 & 65.1 & 56.8 & 60.7 & 44.1 \\
    Qwen3-VL-Plus & \ding{51} & 67.0 & 54.6 & 56.9 & 45.9 & 42.0 & \cellcolor{bestclosed}\textbf{66.7} & 49.3 & 58.9 & 40.9 \\
    Nano-Banana & \ding{55} & 55.4 & 50.2 & 51.8 & 34.5 & 36.6 & 56.7 & 39.4 & 60.4 & 33.2 \\

    \midrule
    \multicolumn{11}{c}{\textit{Open-source MLLMs}} \\
    \midrule
    Qwen-2.5-VL-7B & \ding{55} & 19.5 & 19.0 & 19.2 & 20.6 & 18.7 & 10.7 & 13.9 & 15.0 & 8.9 \\
    Qwen-2.5-VL-32B & \ding{55} & 29.8 & 27.4 & 27.8 & 27.4 & 27.2 & 27.9 & 20.1 & 30.5 & 15.4 \\
    Qwen-2.5-VL-72B & \ding{55} & 30.6 & 19.5 & \cellcolor{bestopen}\textbf{36.4} & 34.5 & 33.5 & 23.9 & \cellcolor{bestopen}\textbf{33.6} & \cellcolor{bestopen}\textbf{48.9} & 21.1 \\
    Gemma-3-27b-it & \ding{55} & 31.3 & 28.4 & 34.4 & 25.8 & 21.0 & 40.0 & 21.0 & 26.9 & 15.8 \\
    InternVL3.5-8B & \ding{55} & 32.3 & \cellcolor{bestopen}\textbf{33.8} & 33.8 & 24.2 & 26.9 & \cellcolor{bestopen}\textbf{43.7} & 16.2 & 14.9 & 16.7 \\
    InternVL3.5-30B & \ding{55} & 22.2 & 19.9 & 15.1 & 24.9 & 24.3 & 22.1 & 17.4 & 18.4 & 11.7 \\
    Keye-VL-1.5-8B & \ding{51} & \cellcolor{bestopen}\textbf{33.1} & 28.0 & 26.2 & 27.0 & 23.6 & 29.5 & 20.9 & 26.3 & 17.1 \\
    \midrule

    \multicolumn{11}{c}{\textit{United MLLMs}} \\
    \midrule
    
    BAGEL-7B-MoT & \ding{55} & 18.1 & 13.1 & 17.1 & 20.8 & 23.0 & 10.9 & 19.4 & 13.3 & 8.3 \\
    BAGEL-Zebra-CoT & \ding{55} & 18.0 & 15.1 & 15.6 & 18.0 & 16.8 & 20.8 & 11.1 & 14.1 & 8.0 \\
    
    MathCanvas-7B & \ding{55} & 29.9 & 27.2 & 17.9 & 40.0 & 35.3 & 23.2 & 29.3 & 40.4 & 21.9 \\
    
    \bottomrule
    \bottomrule
    \end{tabular}
}
\caption{Detailed performance breakdown on the \textbf{GeoAux-Bench-Canvas} subset. Results are cited from \citet{Shi2025MathCanvasIV}. Best scores in \colorbox{bestclosed}{\textbf{closed}} and \colorbox{bestopen}{\textbf{open}} categories are highlighted.}
\label{tab:canvas_results}
\end{table*}

\section{Case Study}
\subsection{Case of Visual Hallucinations}
\label{app:case1}

\begin{figure*}[h]
\centering
\includegraphics[width=0.95\textwidth]{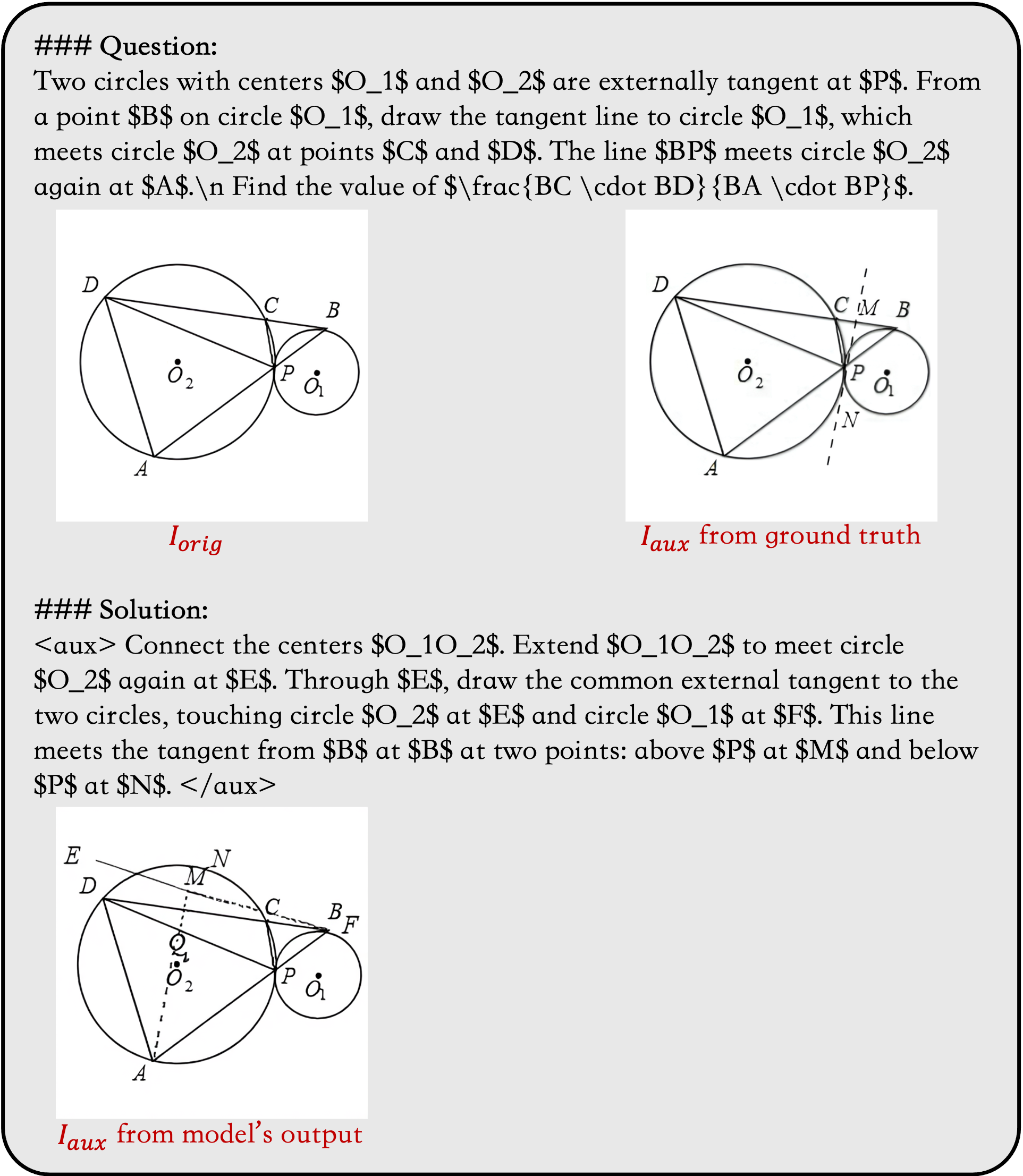}
\caption{Failure Cases of MathCanvas-7B with Visual Hallucinations.}
\label{fig:visual_hallucination}
\end{figure*}

We present representative failure cases of Native Unified MLLMs (e.g., MathCanvas-7B) in Figure \ref{fig:visual_hallucination}, as these models lack the pixel-level control required to produce accurate diagrams. This leads to a Visual-Logic Mismatch: the generated diagram contains severe distortions (e.g., curved lines, missed intersections), causing the model to hallucinate non-existent geometric properties based on the flawed visual feedback, which ultimately derails the reasoning process.

\subsection{Case of the Analytic Shortcut}
\label{app:case2}

We present failure cases demonstrating the model's tendency to bypass geometric intuition in favor of algebraic brute force, termed as \textbf{The Analytic Shortcut}. Instead of employing geometric theorems or auxiliary lines, the model habitually establishes coordinate systems to solve problems via complex equations. As discussed in the main text, while this approach is viable for Senior high school problems, it often proves inefficient or fails to capture the pure geometric logic required for Junior geometry tasks.

\begin{figure*}[h]
\centering
\includegraphics[width=0.95\textwidth]{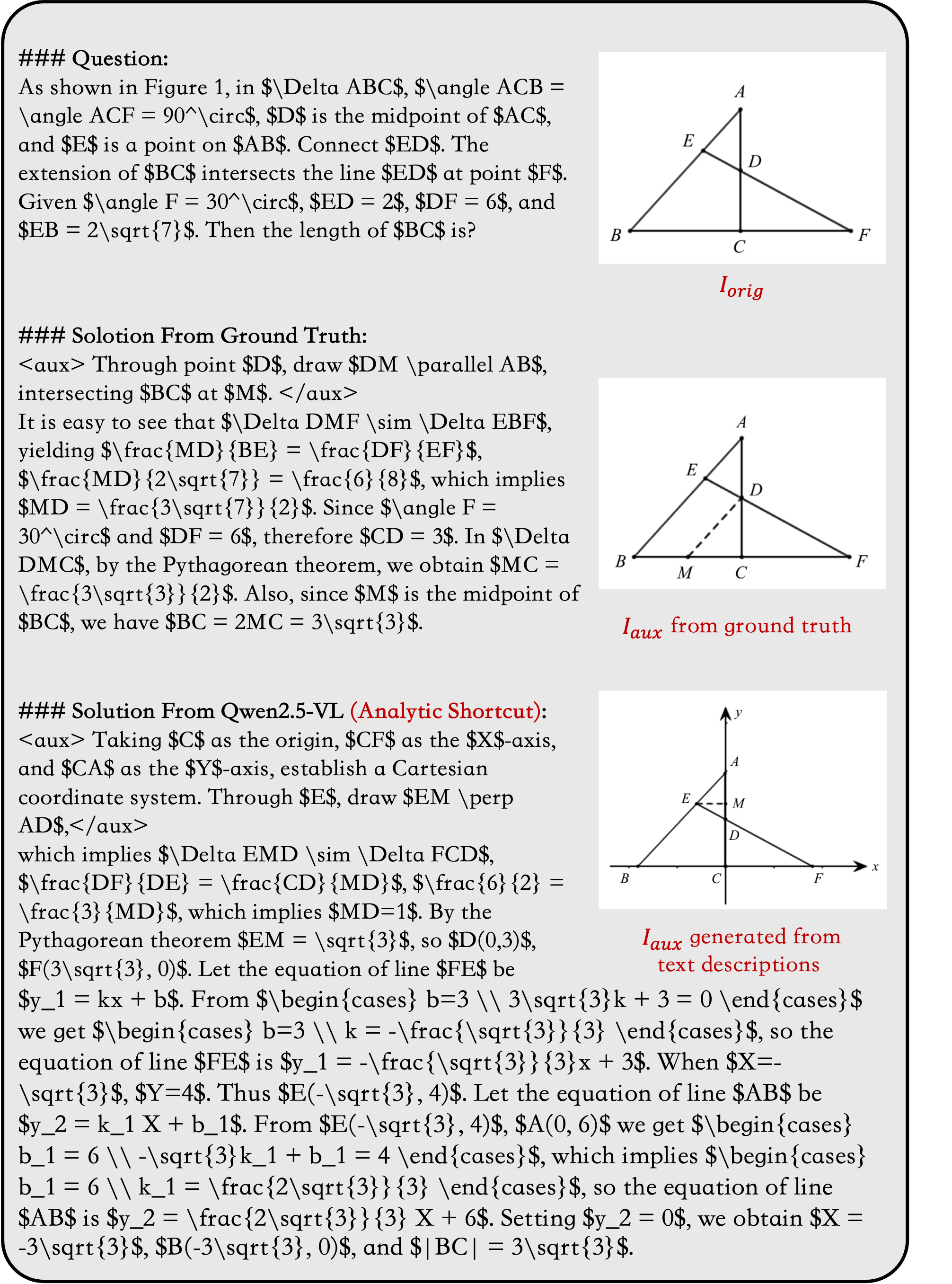} 
\caption{An example where the model forces a coordinate system on a geometry problem, leading to unnecessary complexity.}
\label{fig:analytic_shortcut}
\end{figure*}
\section{Prompt Templates}
\label{app:prompts}

\begin{figure*}[h]
    \centering
    \includegraphics[width=0.95\textwidth]{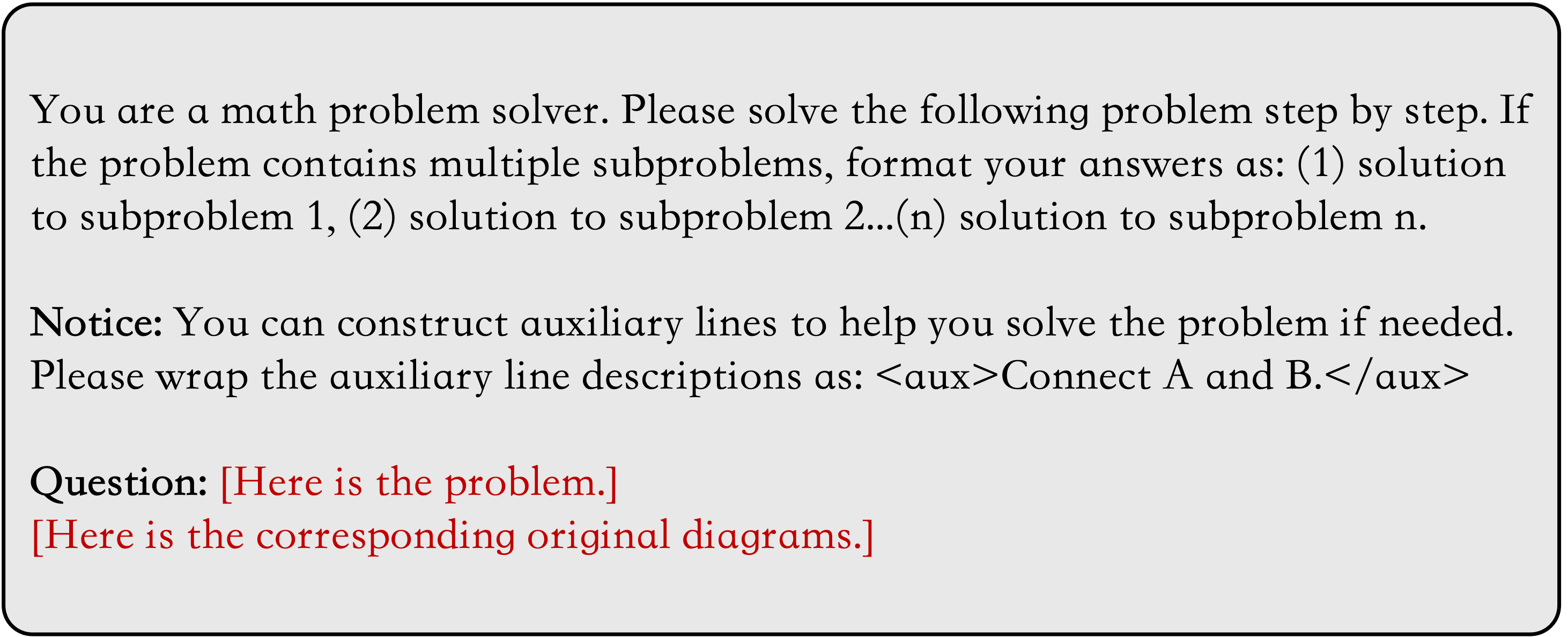} 
    \caption{Standard prompt template (used for $O$, $O^+$ subsets and evaluation).}
    \label{fig:standard_prompt}
\end{figure*}

\begin{figure*}[h]
    \centering
    \includegraphics[width=0.95\textwidth]{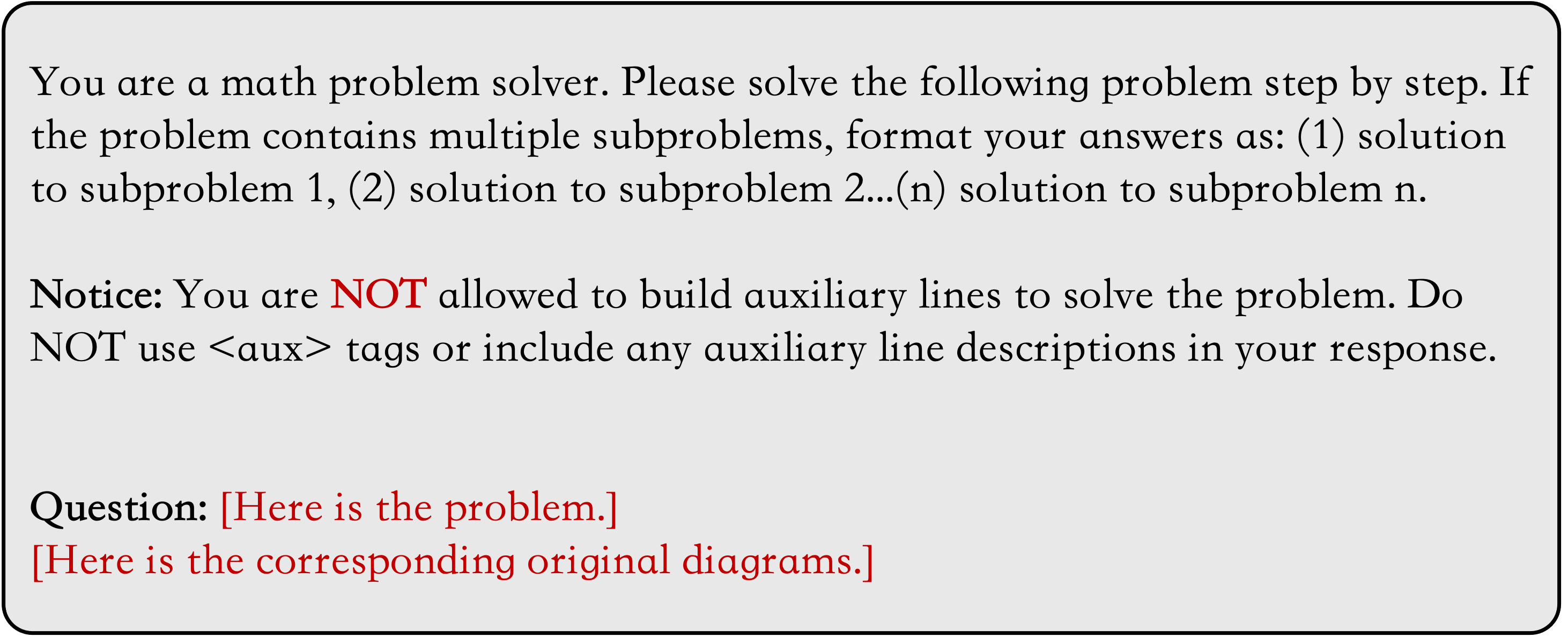} 
    \caption{Prompt template for the Prohibited Subset ($O^-$).}
    \label{fig:prohibited_prompt}
\end{figure*}

\begin{figure*}[h]
    \centering
    \includegraphics[width=0.95\textwidth]{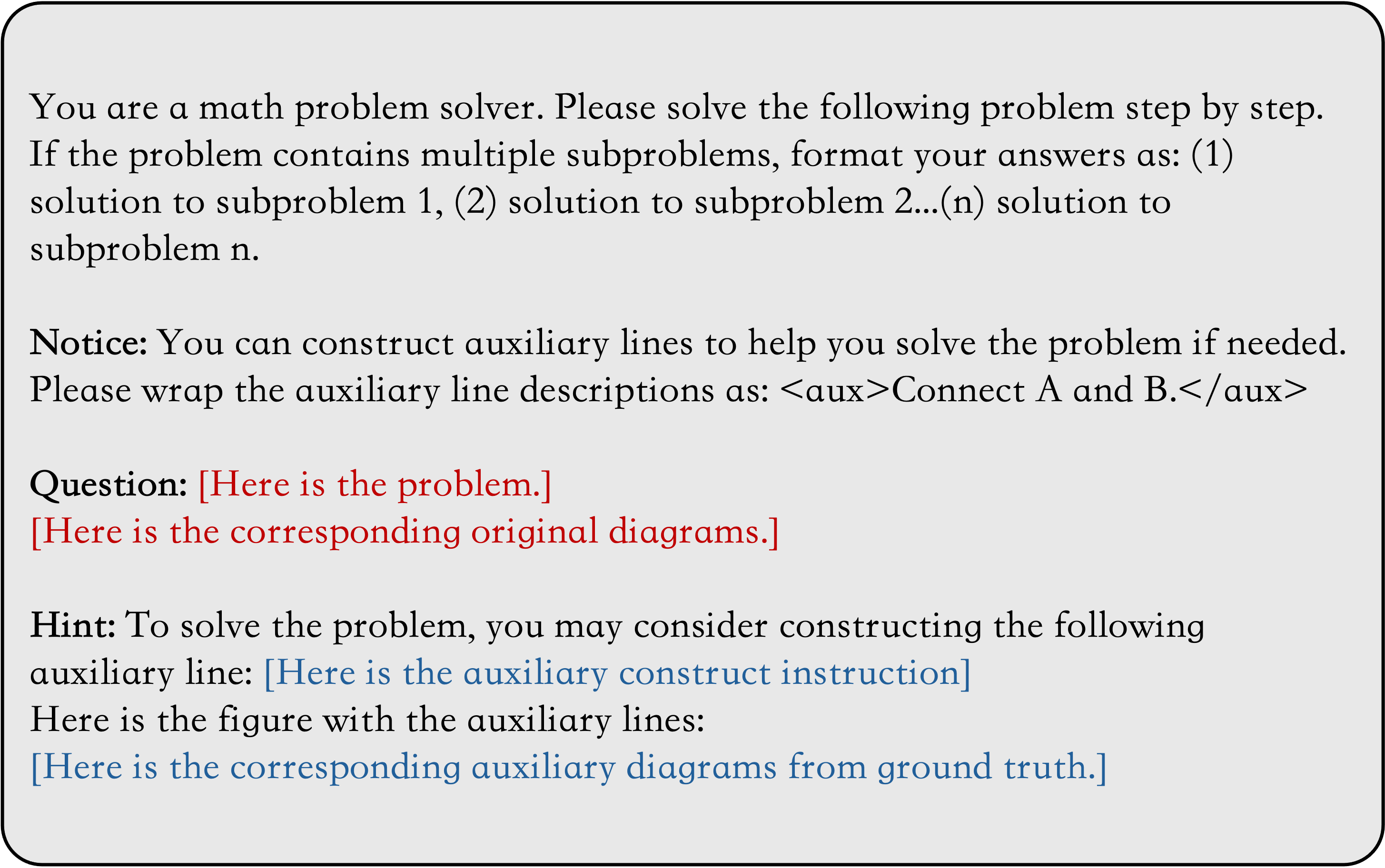}
    \caption{Prompt template for visual-injected re-prompting.}
    \label{fig:re_prompt}
\end{figure*}

\begin{figure*}[h]
    \centering
    \includegraphics[width=0.95\textwidth]{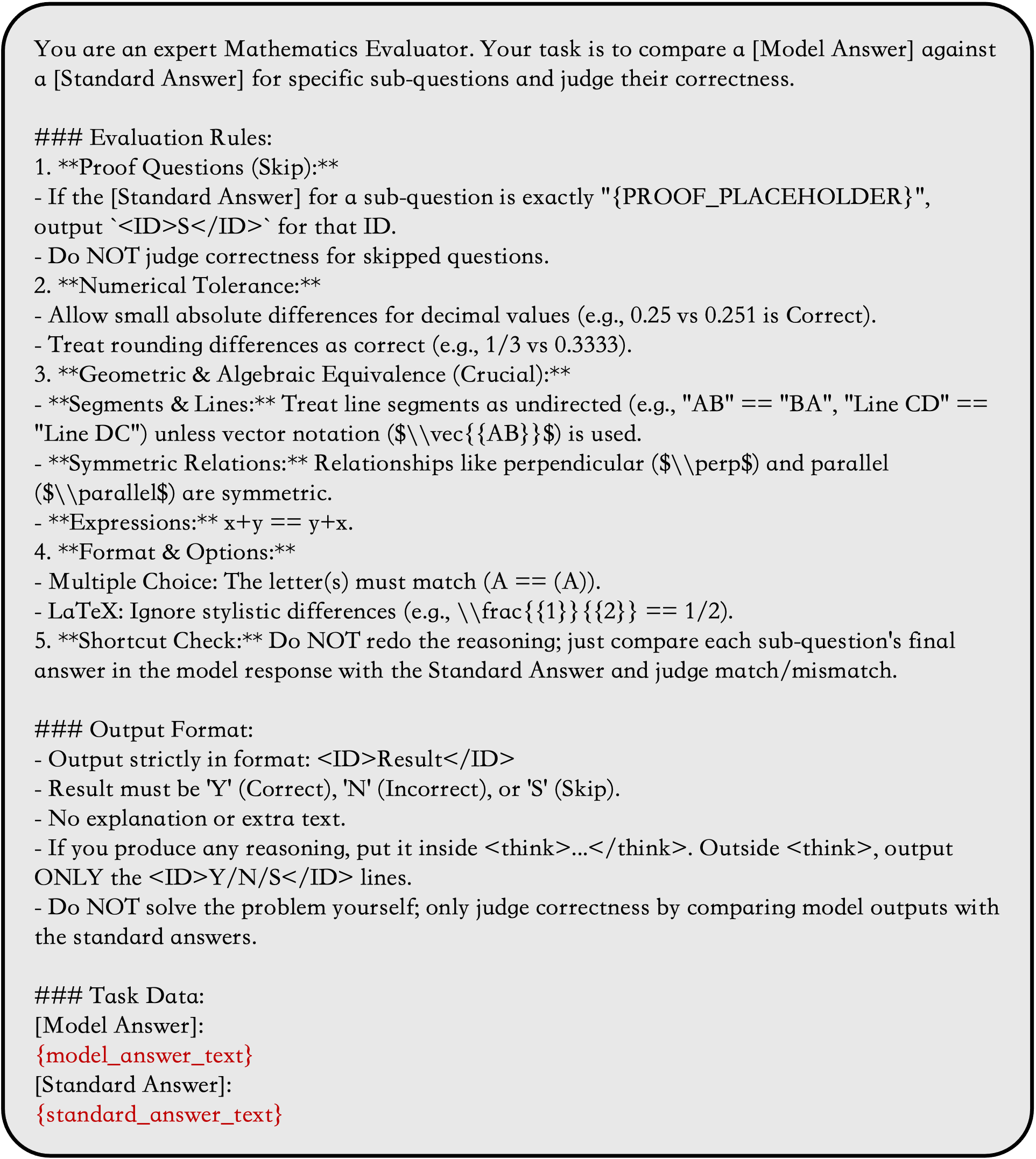} 
    \caption{Prompt template for correctness judgment (used in evaluation and reward calculation).}
    \label{fig:judgeprompt}
\end{figure*}

\begin{figure*}[h]
    \centering
    \includegraphics[width=0.95\textwidth]{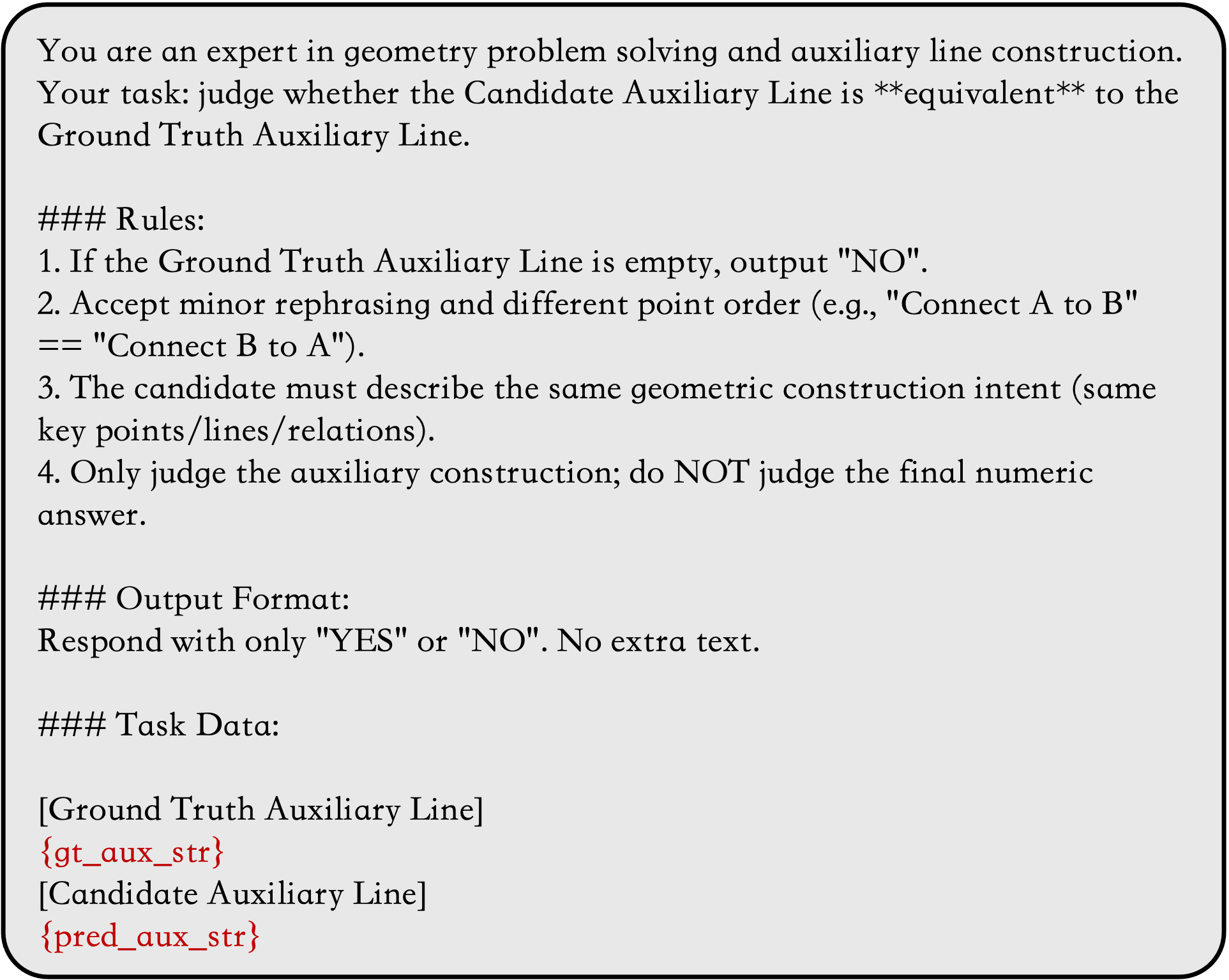}
    \caption{Prompt template for auxiliary construction verification.}
    \label{fig:auxprompt}
\end{figure*}

We present the prompt templates used in Tri-Partition Sampling (Section~\ref{sec:sampling}) and evaluation. Figures~\ref{fig:standard_prompt} and \ref{fig:prohibited_prompt} show the templates for the Natural/Mandatory ($O/O^+$) and Prohibited ($O^-$) subsets, respectively. Figure~\ref{fig:re_prompt} shows the template for visual re-prompting. Figures~\ref{fig:judgeprompt} and \ref{fig:auxprompt} display the prompts for the answer judge and auxiliary verifier.

\section{Implementation Details}
\label{app:hyperparams}

\subsection{Training Hyperparameters}

Table~\ref{tab:hyperparams_training} details the hyperparameters used during the Supervised Fine-Tuning (SFT) warm-up phase and the subsequent A\textsuperscript{2}PO training phase. We utilize \texttt{Qwen2.5-VL-7B-Instruct} as the backbone. For the SFT stage, we freeze the vision tower and the multi-modal projector, updating only the language model. For the A\textsuperscript{2}PO stage, we continue to freeze the vision tower to maintain visual feature stability.

\begin{table}[htbp]
\centering
\small
\caption{\textbf{Training Hyperparameters.} Comparison between the SFT warm-up stage and the A\textsuperscript{2}PO RL stage.}
\label{tab:hyperparams_training}
\begin{tabular}{l c c}
\toprule
\textbf{Hyperparameter} & \textbf{SFT (Warm-up)} & \textbf{A\textsuperscript{2}PO (RL)} \\
\midrule
Base Model & \multicolumn{2}{c}{Qwen2.5-VL-7B-Instruct} \\
Precision & \multicolumn{2}{c}{bfloat16} \\
Optimizer & \multicolumn{2}{c}{AdamW} \\
Learning Rate & 5e-5 & 1e-6 \\
LR Scheduler & Cosine & Constant \\
Warm-up Ratio & 0.1 & 0.0 \\
Global Batch Size & 32 & 24 \\
Gradient Accumulation & 4 & - \\
Max Sequence Length & 8,192 & 8,192 \\
Image Resolution (min/max) & $262,144$ &  $262,144$ \\
Epochs/Steps & 5 epochs & 650 steps \\
\midrule
\multicolumn{3}{l}{\textit{GRPO Specific Settings}} \\
KL Coefficient ($\beta$) & - & 0.01 \\
Rollout Batch Size & - & 72 \\
Generations per Prompt ($N$) & - & 8 \\
\bottomrule
\end{tabular}
\end{table}

\subsection{A\textsuperscript{2}PO Algorithm Coefficients}

Table~\ref{tab:hyperparams_algo} lists the specific weighting coefficients and thresholds defined in the reward shaping mechanism (Section~\ref{sec:rewards}).

\begin{table}[htbp]
\centering
\small
\caption{\textbf{A\textsuperscript{2}PO Reward Coefficients.}}
\label{tab:hyperparams_algo}
\begin{tabular}{l c c}
\toprule
\textbf{Component} & \textbf{Symbol} & \textbf{Value} \\
\midrule
\multicolumn{3}{l}{\textit{Reward Weights}} \\
Accuracy Weight & $w_{acc}$ & 0.70 \\
Format Weight & $w_{fmt}$ & 0.00 \\
Timing Weight & $w_{time}$ & 0.15 \\
Quality Weight & $w_{qual}$ & 0.15 \\
\midrule
\multicolumn{3}{l}{\textit{Thresholds \& Margins}} \\
Timing Significance & $\tau$ & 0.15 \\
PPL Tolerance & $\delta$ & 0.01 \\
\bottomrule
\end{tabular}
\end{table}

\subsection{Inference and Sampling Configuration}

Table~\ref{tab:hyperparams_inference} details the generation parameters. We utilize a strong external model as the \textbf{Aux Verifier} and \textbf{Judge} to ensure the quality of the retrieved auxiliary diagrams and the final answer correctness.

\begin{table}[htbp]
\centering
\small
\caption{\textbf{Generation \& Verifier Configurations.} Settings for training rollouts, the auxiliary verifier, and final evaluation.}
\label{tab:hyperparams_inference}
\begin{tabular}{l c c}
\toprule
\textbf{Parameter} & \textbf{Training Rollout} & \textbf{Evaluation} \\
\midrule
Temperature & 1.0 & 0.0 (Greedy)\\
Top-p & 1.0 & 0.0 \\
Repetition Penalty & 1.05 & 1.08 \\
Max New Tokens & 8,192 & 8,192 \\
\midrule
\multicolumn{3}{l}{\textit{Accuracy Reward Model}} \\
Verifier Model & \multicolumn{2}{c}{Qwen3-30B-A3B-Thinking-2507} \\
Verifier Temperature & \multicolumn{2}{c}{0.0 (Greedy)} \\
Max Tokens & \multicolumn{2}{c}{24576} \\
\midrule
\multicolumn{3}{l}{\textit{Aux Verifier}} \\
Verifier Model & \multicolumn{2}{c}{Qwen3-30B-A3B-Thinking-2507} \\
Verifier Temperature & \multicolumn{2}{c}{0.0 (Greedy)} \\
Max Tokens & \multicolumn{2}{c}{8192} \\
\bottomrule
\end{tabular}
\end{table}
\subsection{Dataset Composition and Splits}
\label{app:data_splits}

To ensure rigorous evaluation, we strictly enforce decontamination between training and evaluation sets. Table~\ref{tab:data_statistics} summarizes the data sources and statistics across the SFT warm-up, A\textsuperscript{2}PO training, and evaluation phases.

\paragraph{Training Data Construction.}
The SFT dataset is constructed using a multi-constraint instruction tuning strategy to initialize controllability. We augment training samples by creating diverse prompt-response pairs, specifically: (1) \textbf{Standard Prompts} paired with solutions containing auxiliary constructions, and (2) \textbf{Prohibited Prompts} paired with solutions strictly devoid of auxiliary lines.
For the A\textsuperscript{2}PO stage, we employ a marginal solvability filtering strategy. We perform 10 inference rollouts per problem using the base model. We retain only samples that exhibit mixed outcomes (i.e., containing both correct and incorrect responses), explicitly discarding trivial (100\% correct) or impossible (0\% correct) instances to maximize gradient efficiency.

\paragraph{Evaluation Data.}
All evaluation results reported in the main paper are based on held-out test sets. For \textbf{GeoAux-Bench}, we reserve a fixed test split that is strictly excluded from all training stages.

\begin{table}[htbp]
\centering
\small
\caption{\textbf{Data Statistics and Splits.} Detailed breakdown of sample counts. Note that for SFT and RL phases, we utilize subsets from the training splits of external datasets to avoid contamination.}
\label{tab:data_statistics}
\begin{tabular}{l l r}
\toprule
\textbf{Stage} & \textbf{Data Source / Subset} & \textbf{Count} \\
\midrule
\multicolumn{3}{l}{\textit{Phase 1: SFT Warm-up (Mixed-Prompt)}} \\
& GeoAux-Bench (Train Split) & 1,000 \\   
& GeomVerse (Train Subset) & 300 \\
& Geometry3k (Train Subset) & 300 \\
\midrule
\multicolumn{3}{l}{\textit{Phase 2: A\textsuperscript{2}PO Training (Solvability Filtered)}} \\
& GeoAux-Bench (Train Split) & 1,500 \\   
& GeomVerse (Train Subset) & 300 \\
& Geometry3k (Train Subset) & 300 \\
\midrule
\multicolumn{3}{l}{\textit{Phase 3: Evaluation (Held-out Test Sets)}} \\
& \textbf{GeoAux-Bench (Test)} & \textbf{1,217} \\ 
& GeomVerse (Test) & 1,000 \\
& Geometry3k (Test) & 901 \\
\bottomrule
\end{tabular}
\end{table}

\end{document}